\documentclass{article}

 \usepackage[preprint]{neurips_2026}

\usepackage[utf8]{inputenc}
\usepackage[T1]{fontenc}
\usepackage{hyperref}
\usepackage{url}
\usepackage{booktabs}
\usepackage{amsfonts}
\usepackage{nicefrac}
\usepackage[protrusion=true,expansion=true,activate={true,nocompatibility},final,tracking=true,kerning=true,spacing=true,disable=footnote]{microtype}
\usepackage{xcolor}

\PassOptionsToPackage{numbers}{natbib}
\usepackage{graphicx,subcaption}
\usepackage{enumitem}
\usepackage{algorithm}
\usepackage{algorithmic}
\usepackage{booktabs}
\usepackage{ragged2e}
\usepackage{wrapfig}
\usepackage{arydshln}
\usepackage{setspace}
\usepackage{adjustbox}
\usepackage{framed}
\usepackage{wrapfig}

\usepackage{pgfplots}
\usepackage{pgfplotstable}
\pgfplotsset{compat=1.18}
\usepgfplotslibrary{groupplots}

\usepackage{amssymb}
\usepackage{mathtools}
\usepackage{amsthm}
\usepackage{multirow}
\usepackage{xparse}

\usepackage{pgfplots}
\usepgfplotslibrary{polar,groupplots}
\pgfplotsset{compat=1.18}

\usepackage{pgfplots}
\usepackage{subcaption}
\usepackage{xcolor}

\usepackage{empheq}
\usepackage[most]{tcolorbox}
\usepackage{calc}

\newtcbox{\eqbox}{
  colback=blue!5, colframe=blue!5, boxrule=0pt, arc=3pt,
  nobeforeafter, math upper, tcbox raise base,
  enhanced, boxsep=2pt, left=0pt, right=0pt, top=0pt, bottom=0pt
}

\pgfplotsset{compat=1.18}

\usepackage{pgfplots}
\usepackage{subcaption}
\usepackage{xcolor}
\usepackage{pgfplotstable}
\pgfplotsset{compat=1.18}

\definecolor{m1}{RGB}{160,160,160}
\definecolor{m2}{RGB}{130,146,166}
\definecolor{m3}{RGB}{97,148,147}
\definecolor{m4}{RGB}{92,124,189}
\definecolor{m5}{RGB}{38,69,120}
\definecolor{m6}{RGB}{180,120,90}
\definecolor{m7}{RGB}{120,170,110}
\definecolor{m8}{RGB}{150,110,170}

\usepackage{xparse}
\newlength{\meanw}
\newlength{\stdw}

\theoremstyle{plain}

\theoremstyle{remark}

\hypersetup{hidelinks}

\definecolor{titlebg}{HTML}{E6E5EF}  

\definecolor{linkblue}{HTML}{116E8A}

\newcommand{\mitlogoheight}{0.85cm}
\newcommand{\kuleuvenlogoheight}{1.33cm}
\newcommand{\logorowheight}{1.33cm}

\makeatletter
\newcommand{\fairauthors}[1]{\gdef\@fairauthors{#1}}
\newcommand{\fairaffiliations}[1]{\gdef\@fairaffiliations{#1}}
\newcommand{\fairabstract}[1]{\gdef\@fairabstract{#1}}
\newcommand{\faircorrespondence}[1]{\gdef\@faircorrespondence{#1}}
\gdef\@fairauthors{}\gdef\@fairaffiliations{}
\gdef\@fairabstract{}\gdef\@faircorrespondence{}

\newcommand{\synibmaketitle}{%
  \thispagestyle{empty}%
  \begingroup
  \tcbset{enhanced,frame hidden,
          left=0.7cm,right=0.7cm,top=0.32cm,bottom=0.28cm,
          arc=5pt,colback=titlebg,
          before skip=0pt,after skip=0.3cm,
          grow to left by=1.5pt,grow to right by=1.5pt}
  \begin{tcolorbox}
    \setlength{\parindent}{0pt}%
    \raggedright
    {\Large\bfseries \@title\par}
    \vskip 0.16cm
    {\small\bfseries \@fairauthors\par}
    \vskip 0.06cm
    {\small \@fairaffiliations\par}
    \vskip 0.16cm
    {\normalsize \@fairabstract\par}
    \vskip 0.16cm
    \noindent
    \begin{minipage}[c]{0.48\linewidth}
      \raggedright
      \ifx\@faircorrespondence\@empty\else
        {\small\textbf{Correspondence:} \@faircorrespondence}%
      \fi
    \end{minipage}%
    \hfill
    \begin{minipage}[c]{0.50\linewidth}
      \raggedleft
      \mbox{%
        \parbox[c][\logorowheight][c]{\widthof{\includegraphics[height=\mitlogoheight]{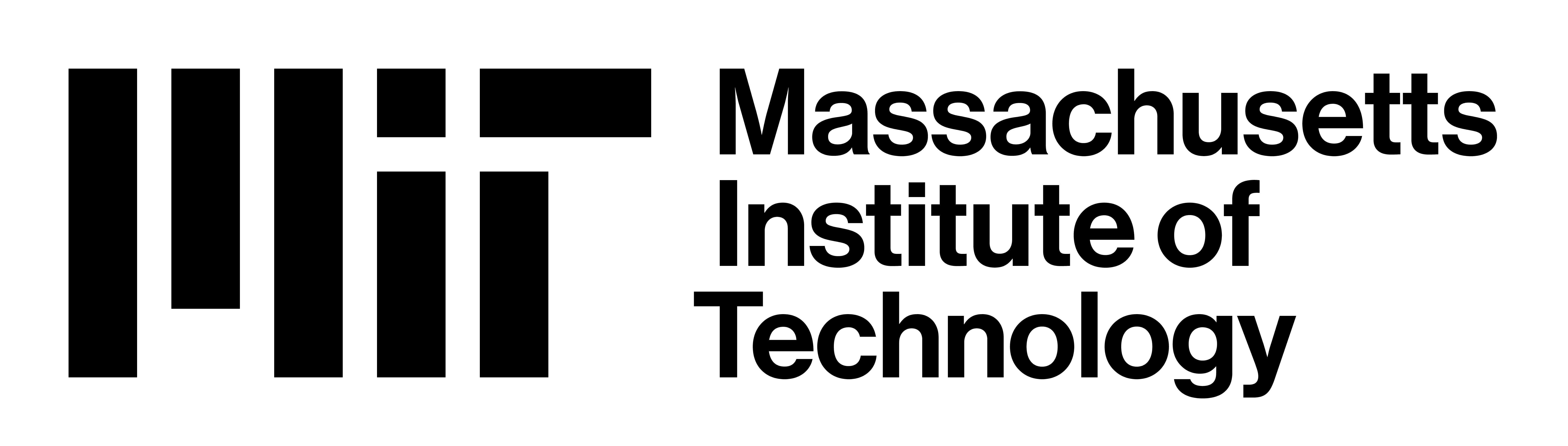}}}{%
          \centering\includegraphics[height=\mitlogoheight]{Figures/logos/mit_lockup_std-three-line_rgb_black.png}%
        }%
        \hspace{0.4cm}%
        \parbox[c][\logorowheight][c]{\widthof{\includegraphics[height=\kuleuvenlogoheight]{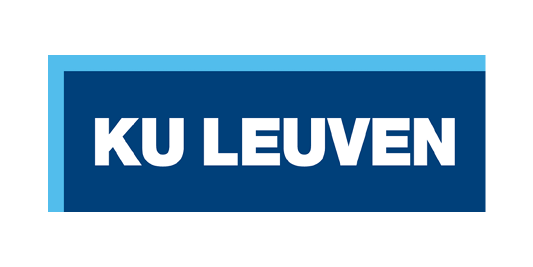}}}{%
          \centering\includegraphics[height=\kuleuvenlogoheight]{Figures/logos/KU-Leuven-logo.png}%
        }%
      }%
    \end{minipage}%
  \end{tcolorbox}
  \endgroup
}
\makeatother

\title{SynIB: Informational Bottleneck for Maximizing Synergy in Multimodal Learning}
\author{}

\fairauthors{%
  Konstantinos Kontras$^{1,2}$,
  Teodora Gagaleska$^{1}$,
  Thomas Strypsteen$^{1}$,
  Christos Chatzichristos$^{1}$,
  Matthew Blaschko$^{1}$,
  Maarten De Vos$^{1,\dagger}$,
  Paul Pu Liang$^{2,\dagger}$%
}
\fairaffiliations{%
  $^{1}$KU Leuven,\quad
  $^{2}$MIT,\quad
  $^{\dagger}$Equal supervision%
}
\faircorrespondence{\href{mailto:kkontras@mit.edu}{\texttt{kkontras@mit.edu}}}

\fairabstract{%
A central objective in multimodal learning is to capture synergy: task-relevant information that arises only from the joint use of multiple modalities, and is not available from any single modality alone. While most approaches operate at the architectural level through larger or more complex fusion models, we propose a complementary axis: shaping the training objective itself. Standard training often emphasizes unimodal or redundant information, falling short on examples that require cross-modal reasoning. We formalize multimodal synergy through information theory and introduce the Synergistic Information Bottleneck (SynIB), a scalable objective that targets synergy directly. To prioritize learning synergy, SynIB motivates the model to predict accurately from all modalities while penalizing confidence when information from any modality is withheld. Alongside the standard task loss, the model runs forward passes with one modality masked at a time and is penalized for remaining confident, which would indicate reliance on unimodal cues rather than cross-modal interactions. We validate SynIB in two regimes. On synthetic XOR tasks where the ground-truth synergy is known by construction, standard training fails to recover it while SynIB does. On five real-world benchmarks, including three MultiBench affective tasks, Hateful Memes with CLIP-ViT and DeBERTa backbones, and a controllable irony extension of CREMA-D we introduce, SynIB improves accuracy on synergy-dependent examples by up to 7.8\% and overall accuracy by up to 3.8\%.%
}

\begin{document}

\synibmaketitle

\section{Introduction}

Multimodal learning aims to combine information from multiple sources to improve prediction. A central challenge is capturing synergistic information: predictive signal that arises from interactions between modalities rather than from any modality alone~\citep{liang2024foundations}. In unimodal learning, nonlinear activations enable feature interactions and techniques such as weight decay~\citep{krogh1991simple} and dropout~\citep{srivastava2014dropout} reduce reliance on individual features~\citep{goodfellow2016deep}; analogous mechanisms for cross-modal interactions remain underexplored.

Multimodal models sometimes over-rely on a single modality during training, favoring signals that yield faster optimization progress, a phenomenon known as multimodal competition~\citep{huang2022modality}. Numerous methods address this by estimating modality contributions and rebalancing learning through gradient modulation, auxiliary losses, or related interventions, but yield limited gains on synergistic cases~\citep{MCR}, indicating that correcting modality imbalance alone is insufficient to induce cross-modal synergy. A complementary line captures cross-modal interactions through architectural design, from fusion mechanisms~\citep{zadeh2017tensor, tsai2019multimodal} to masked multimodal pretraining~\citep{singh2022flava}, expanding capacity for interaction without directly targeting synergistic prediction. Representation-level objectives such as contrastive alignment~\citep{radford2021learning} do shape training, but pull modalities toward agreement, a property closer to redundancy than to synergy. What is missing across these lines is a training signal that directly prioritizes synergistic prediction on the downstream task. Our approach fills this gap: it targets synergy at the loss level, is orthogonal to architectural and pretraining choices, and is complementary to rebalancing methods, which address a related but distinct failure mode.

To this end, we formalize multimodal synergy through an information-theoretic lens and introduce the \emph{Synergistic Information Bottleneck} (SynIB). The idea follows from the definition: synergistic information is the information that vanishes when any single modality is removed. A model that relies on synergy must therefore become uncertain when one modality is masked; a model that remains confident is relying on unimodal or redundant cues. SynIB turns this observation into a training signal by penalizing confident predictions under modality-wise masking, pressuring the model toward cross-modal dependencies on the examples where they matter most. Our main contributions are:
\begin{enumerate}[leftmargin=0.3cm, labelsep=0.1cm, itemsep=0.0cm]
    \item SynIB, a training objective that targets synergy by penalizing confident predictions under modality-wise masking, requiring no prior knowledge of whether synergy exists in the data.
    \item A motivating analysis showing that the failure to learn synergy is not caused by interference from unimodal signals; rather, synergistic cues are learned but quickly overfit, limiting generalization.

    \item Extensive evaluation across real-world benchmarks from MultiBench and HatefulMemes, including pretrained CLIP-ViT and DeBERTa backbones, a new irony extension of CREMA-D, and controlled Gaussian XOR tasks. On these SynIB improves synergy-dependent test set accuracy by up to 7.8\% and overall accuracy by up to 3.8\%.
\end{enumerate}

\section{Related Work}
\paragraph{Partial Information Decomposition (PID).} PID formalizes synergy, redundancy, and unique information across multiple sources with respect to a target~\citep{williams2010nonnegative, griffith2014quantifying, bertschinger2014quantifying}. In multimodal learning, it has primarily served as a post-hoc analytical tool to quantify information distribution in trained models~\citep{liang2023quantifying}, where synergy is operationalized as the predictive gap between a fusion model and an ensemble of unimodal models. Recent work incorporates PID-inspired quantities into training objectives via contrastive factorization~\citep{liang2024factorized, wen2025infmasking} and interaction-aware architectures~\citep{xin2025i2moe}. While effective for disentanglement, attribution, and interpretability, these approaches treat synergy as a structural property to be isolated rather than as a learning optimization signal.

\paragraph{Multimodal representation learning.} A broad field of works captures cross-modal interactions through architectural and representational design: fusion mechanisms from tensor fusion~\citep{zadeh2017tensor} to cross-attention transformers~\citep{tsai2019multimodal, lu2019vilbert}, alignment objectives such as multimodal contrastive learning~\citep{radford2021learning}, masked multimodal pretraining~\citep{singh2022flava, baevski2022data2vec}, and factorized representations into modality-invariant and modality-specific subspaces~\citep{liang2024factorized}. These approaches expand the capacity for cross-modal interaction but none directly shape the training signal on the downstream task to penalize models that ignore synergistic cues, leaving even expressive architectures free to settle into unimodal or redundant solutions when those minimize loss faster~\citep{huang2022modality}. Our objective is complementary: it operates at the loss level and can be combined with any of these architectural choices.

\paragraph{Multimodal competition.} A complementary line of work addresses cross-modal learning failures as multimodal competition, estimating modality contributions and rebalancing learning accordingly. Strategies include measuring contribution via unimodal performance~\citep{MSLR, OGMGE, vielzeuf2018centralnet, MLB, PMR} and enforcing balance through gradient modulation or decomposition~\citep{MMPareto}, reinitializing modality-specific components to escape suboptimal trajectories~\citep{wei2024diagnosing}, perturbation-based attribution such as Shapley approximations~\citep{AGM}, permutation importance~\citep{MCR}, counterfactual modality removal~\citep{ji2022increasing}, and ensemble-based strategies that sidestep fusion~\citep{ReconBoost}. While effective at preventing modality collapse, these methods assume balanced contributions suffice for multimodal learning. As MCR~\citep{MCR} shows, however, correcting modality imbalance does not by itself induce synergy.


\vspace{-10pt}
\section{Method}

\paragraph{Problem setup.} 
We consider a supervised multimodal prediction task with two input modalities $X_1$ and $X_2$ and a target $Y$: $(X_1 \in \mathbb{R}^{d_1},\, X_2 \in \mathbb{R}^{d_2},\, Y \in \mathcal{Y}) \sim p(x_1,x_2,y)$.\footnote{We present the formulation for two modalities for clarity; the SynIB objective 
is symmetric across modalities and extends to $n>2$ settings by adding counterfactual passes on the subset of modalities.} Modality-specific encoders $f_{\theta_i} : \mathbb{R}^{d_i} \to \mathbb{R}^{m_i}$ produce $Z_i = f_{\theta_i}(X_i)$ for $i \in \{1,2\}$, a fusion network $f_{\theta_{12}} : \mathbb{R}^{m_1}\times\mathbb{R}^{m_2}\to\Delta(\mathcal{Y})$ it's multimodal outputs $\hat{Y}_{12} = f_{\theta_{c_{12}}}(Z_1, Z_2)$, and the unimodal encoders $f_{\theta_{c_{i}}} : \mathbb{R}^{m_i}\times\mathbb{R}$ the unimodal outputs $\hat{Y}_{i} = f_{\theta_{c_{i}}}(Z_i)$. We collect all trainable parameters as $\theta = \{\theta_1, \theta_2\, \theta_{c_{1}}, \theta_{c_{2}}, \theta_{c_{12}}\}$.

\paragraph{Synergy as a learning goal.} Partial Information Decomposition~\citep{williams2010nonnegative, bertschinger2014quantifying} decomposes the predictive mutual information into four atoms:
\begin{equation}
I(Y; X_1, X_2) \;=\; R(Y; X_1, X_2) \;+\; U_1(Y; X_1 \mid X_2) \;+\; U_2(Y; X_2 \mid X_1) \;+\; S(Y; X_1, X_2),
\label{eq:pid}
\end{equation}
where $R$ is redundant across modalities, $U_i$ is unique to $X_i$, and $S$ is accessible only from the joint $(X_1, X_2)$. Following~\citet{bertschinger2014quantifying}, synergy is the gap between $I_p(Y; X_1, X_2)$ and its minimum over distributions preserving the bivariate marginals $(Y, X_1)$ and $(Y, X_2)$:
\begin{equation}
S(Y; X_1, X_2) = \underbrace{I_p(Y; X_1, X_2)}_{\substack{\text{predictive information} \\ \text{in the joint distribution}}} - \underbrace{\min_{q \in \Delta_p} \int q(y, x_1, x_2) \log \frac{q(y \mid x_1, x_2)}{q(y)} \, dy\, dx_1\, dx_2}_{\text{maximum predictive information from the unimodal marginals}},
\label{eq:synergy}
\end{equation}
with $\Delta_p = \{\, q : q(y, x_i) = p(y, x_i),\ i \in \{1,2\}\,\}$. The first term is the predictive information available under the joint distribution; the second upper-bounds what any predictor whose dependence on $(X_1, X_2)$ factors through the bivariate marginals. Standard cross-entropy training maximizes $I(Y; X_1, X_2)$ as a single scalar but offers no mechanism to allocate gradient signal to $S$ rather than $R$, $U_1$, or $U_2$~\citep{liang2023quantifying}, motivating an objective that targets $S$ directly.


\subsection{A Motivating Analysis: What Makes Synergy Hard to Learn?}
\label{sec:failure_mode}
A common hypothesis attributes the difficulty of learning synergy to gradient interference between modalities or PID sources~\citep{MMPareto}.We probe this on a
bimodal XOR with single-source examples (App.~\ref{app:pid_xor}),
attributing gradient signal to specific components.

\begin{figure}
      \centering
      \includegraphics[width=1\linewidth]{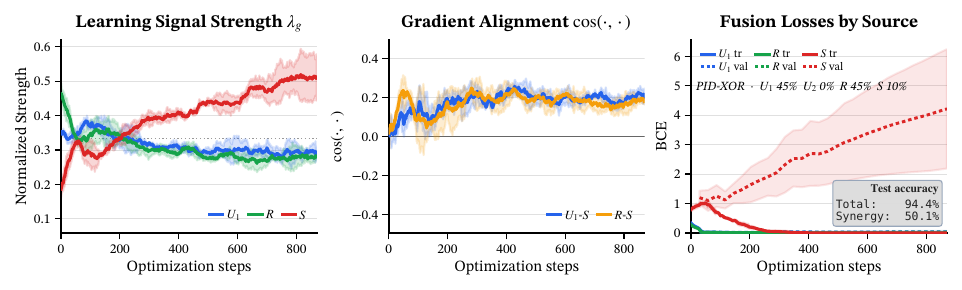}
    \caption{\textbf{Gradient geometry across PID sources
   under vanilla fusion.}                              
  A model is trained on examples drawn from the three  
  PID sources, $U_1$, $R$, and $S$, with $U_2 = 0$ by  
  construction (details in Sec.~\ref{sec:synthetic}).
  \textbf{Left:} Per-group learning signal strength is 
  substantial for all sources, meaning that the examples of that source create gradient capable of changing the parameters, with synergistic      
  examples producing the largest $\lambda_g$.
  \textbf{Center:} Gradient alignment between source
  pairs stays near zero or slightly positive throughout training, indicating
   no destructive competition. \textbf{Right:} Fusion
  BCE by source on intact inputs (solid train, dashed
  validation). Synergistic training loss drops to zero
  while validation rises, indicating overfitting on the
   scarce synergistic examples and leaving test synergy
   accuracy at chance ($50.1\%$; total $94.4\%$).}      \label{fig:pid_ntk_geometry}
  \end{figure}

\paragraph{Observations.} We diagnose vanilla training using two per-source quantities derived from the empirical Neural Tangent Kernel (NTK)~\citep{jacot2018neural, chizat2019lazy}, which characterizes training dynamics through inner products of per-example gradients: the \emph{learning signal strength} $\lambda_g$, measuring how strongly examples from PID source $g$ drive parameter updates, and the \emph{gradient alignment} $\cos(g,h)$, measuring interference between sources (derivations in App.~\ref{app:ntk_pid}). Figure~\ref{fig:pid_ntk_geometry} shows synergistic examples receiving the largest $\lambda_g$ (left) and pairwise alignments staying near zero (center): synergy is neither starved of gradient signal nor in destructive competition with other sources. Yet its training loss drops while validation loss rises (right), indicating overfitting on a
scarce subset rather than inability to learn. This localizes the
failure to training dynamics rather than representational capacity.
SynIB is designed for this regime, motivating the model to learn from 
cross-modal predictions throughout training; App.~\ref{app:ntk_synib}
reports the diagnostics under SynIB.


\subsection{The SynIB Training Objective}

The above analysis suggests synergy fails because the model converges
to unimodal or redundant shortcuts before synergistic cues can be
learned~\citep{huang2022modality}. SynIB addresses this by penalizing
the model when it remains confident under a modality corruption
$\tilde{X}_1$ (with latent $\tilde{Z}_1 = f_{\theta_1}(\tilde{X}_1)$):
confidence under the counterfactual $(\tilde{X}_1, X_2)$ indicates the
model reached its answer without cross-modal interaction, pushing it
toward joint use of both modalities. The objective combines
cross-entropy on intact inputs with a confidence penalty on
counterfactuals.

The remainder of this section develops each component. Sec.~\ref{sec:mipd} grounds the confidence penalty in information-theory; Sec.~\ref{sec:variational} derives a tractable variational surrogate; Sec.~\ref{sec:masking} specifies the corruption operator $\tilde{X}_1$; and Sec.~\ref{sec:objective} assembles the full objective.

\subsection{Quantifying Complementarity from Counterfactual Predictions}
\label{sec:mipd}
The confidence penalty under corruption measures how much the model's predictions change when information from one modality is removed. This quantity has a principled interpretation, derived from the information $X_1$ contributes to predicting $Y$ beyond what is available from $X_2$ alone.

The information $X_1$ carries about $Y$ given $X_2$ is captured by the conditional mutual information $I(X_1;Y\mid X_2) = H(Y\mid X_2) - H(Y\mid X_1,X_2)$. Computing this quantity requires marginalizing $X_1$ out of $H(Y \mid X_2)$, which is intractable in high dimensions. We replace this marginalization with a single targeted perturbation $\tilde{X}_1$ designed to remove $X_1$'s task-relevant content, and measure the resulting loss in conditional predictability, the \emph{Mutual Information Perturbation Difference}:
\begin{equation}
\mathrm{MIPD}_1 = H(Y \mid \tilde{X}_1, X_2) - H(Y \mid X_1, X_2) = I(X_1; Y \mid X_2) - I(\tilde{X}_1; Y \mid X_2) \leq I(X_1; Y \mid X_2).
\label{eq:mipd-def}
\end{equation}
Maximizing $\mathrm{MIPD}_1$ maximizes a lower bound on the contribution of $X_1$ beyond $X_2$; derivation and tightness conditions are in App.~\ref{app:mipd-bounds}. This transforms the question of estimating CMI into the design problem of constructing $\tilde{X}_1$, addressed in Sec.~\ref{sec:masking}.

\begin{figure*}[t]
    \centering
    \includegraphics[width=0.98\linewidth]{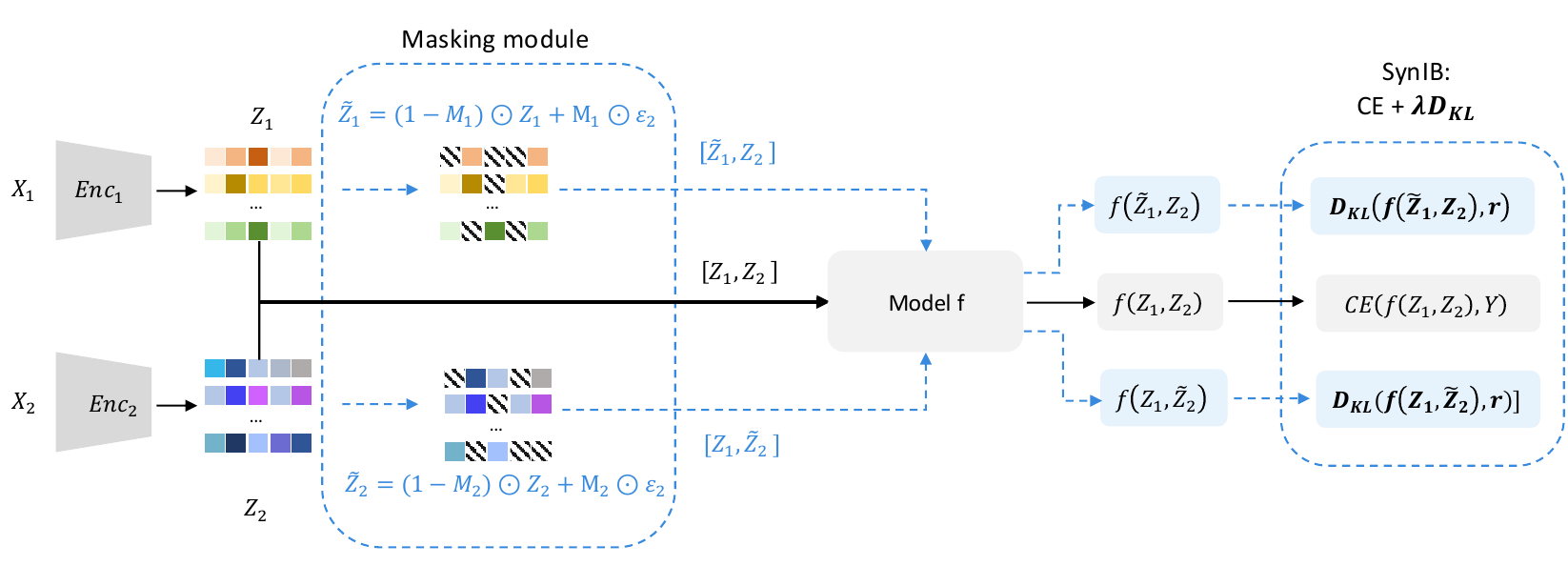}
    \caption{\textbf{SynIB overview.} Standard multimodal fusion (black) trains a model to predict $Y$ from $(Z_1, Z_2)$, leaving optimization free to settle on unimodal or redundant cues. SynIB (blue) adds counterfactual passes in which one modality is replaced with a feature-masked version $\tilde Z_i$ that removes its task-relevant content, and penalizes the model when its predictions remain confident under this corruption. Confidence under masking signals reliance on a single modality; the KL penalty against a reference distribution $r$ pushes the model toward predictions that depend on the joint $(Z_1, Z_2)$, so the cross-entropy objective prioritizes learning from cross-modal interactions.}
    \label{fig:method}
\end{figure*}


\subsection{Variational Approximation for Tractable Cross-Modal Optimization}
\label{sec:variational}
Eq.~\eqref{eq:mipd-def} is not directly optimizable: the conditional entropies require access to the true predictive distributions $p(y \mid x_1, x_2)$ and $p(y \mid \tilde{x}_1, x_2)$. We obtain a tractable surrogate by approximating each entropy through variational quantities the model can compute.

For $H(Y \mid X_1, X_2)$, a variational decoder $q_\theta(y \mid x_1, x_2)$ and the non-negativity of KL~\citep{barber2004algorithm} give the standard upper bound $H(Y \mid X_1, X_2) \le \mathbb{E}_{p(x_1,x_2,y)}[-\log q_\theta(y \mid x_1, x_2)]$, recovering the cross-entropy loss. For $H(Y \mid \tilde{X}_1, X_2)$, the reference-based entropy identity~\citep{tishby2000information, alemi2016deep} gives
\begin{align}
H(Y\mid \tilde{X}_1,X_2)
=
C_r - \mathbb{E}_{p(\tilde{x}_1,x_2)}\,
D_{\mathrm{KL}}\!\left(p(\cdot\mid \tilde{x}_1,x_2)\,\|\,r(\cdot)\right),
\label{eq:ref-identity}
\end{align}
where $C_r = \mathbb{E}_{p(y)}[-\log r(y)]$ is constant in $\theta$. Substituting $q_\theta$ for $p$ in Eq.~\eqref{eq:ref-identity} defines the variational surrogate $\widehat{H}_\theta(Y\mid \tilde{X}_1, X_2)$, which recovers the true entropy when $q_\theta = p$. Combining $\widehat{H}_\theta$ with the cross-entropy upper bound on $H(Y\mid X_1, X_2)$ gives a tractable surrogate for $\mathrm{MIPD}_1$, which coincides with the true value when the variational family contains $p$. The two terms together form the SynIB training objective, assembled in Sec.~\ref{sec:objective}. Variational families, closed-form KL divergences, and properties of the surrogate are in Appendices~\ref{app:variational} and~\ref{app:mipd-bounds}.

\subsection{Counterfactual Corruption via Feature Masking}
\label{sec:masking}

The MIPD lower bound from Sec.~\ref{sec:mipd} tightens with smaller $I(\tilde{X}_1; Y \mid X_2)$, giving a concrete design criterion: choose $\tilde{X}_1$ to minimize residual task-relevant information while perturbing as little of $X_1$ as possible. We implement the corruption as feature-wise masking through a binary mask $M_1 \in \{0,1\}^{d_1}$ and a gate $\tilde{z}_1 = (1 - M_1) \odot z_1 + M_1 \odot \epsilon$, where $\epsilon$ is noise independent of $Y$ and coordinates with $M_{1,j} = 1$ are corrupted. Feature-level control gives more precise residual minimization than global perturbations, and the operator can be applied equivalently to $x_i$ or the latent $z_i$.

The choice of $M_1$ determines what SynIB ends up penalizing. The ideal choice would corrupt cross-modal cues while leaving unimodal ones intact, so that any remaining confidence under $(\tilde{X}_1, X_2)$ comes from unimodal reliance and gets penalized. Since unimodal cues are not labeled, we identify them adversarially: a \textbf{learned mask} $M_{\psi_1} \in \{0,1\}^{d_1}$ is trained to flag the minimum set of coordinates whose corruption collapses unimodal prediction,
\begin{equation}
\mathcal{L}_{\text{mask}}(\psi_1)
=
-\mathrm{CE}\!\left(f_{\theta_{c_1}}\!\left((1 - M_{\psi_1}) \odot x_1 + M_{\psi_1} \odot \epsilon\right),\, y\right)
+ \lambda_M \|M_{\psi_1}\|_1,
\label{eq:mask-loss}
\end{equation}
and the SynIB corruption mask is its inverse, $M_1 = \mathbf{1} - M_{\psi_1}$. This isolates the features the unimodal predictor relies on: SynIB preserves them and corrupts the rest, leaving the model with only unimodal cues at the counterfactual pass. Eq.~\eqref{eq:mask-loss} is equivalent to finding the sparsest perturbation that maximally degrades the unimodal log-likelihood (App.~\ref{app:density}). 

\begin{figure}[t]
    \centering
    \vspace{-6mm}
    \includegraphics[width=\linewidth]{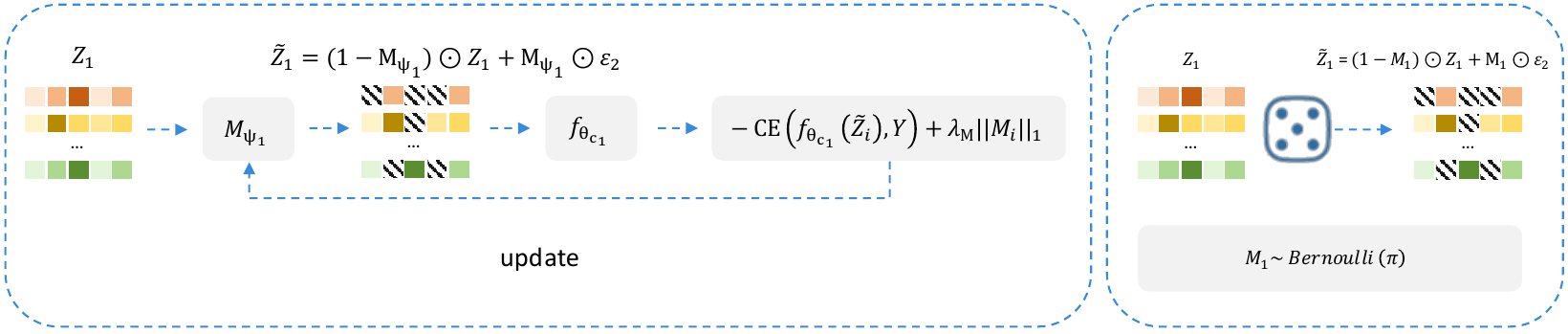}
    \caption{Two strategies for constructing the counterfactual mask $M$. \textbf{Left:} Learned masking trains $M_{\psi_1}$ adversarially against a unimodal predictor with a sparsity penalty, isolating the features needed for unimodal prediction; SynIB inverts this mask so the corruption removes everything else, leaving only unimodal cues intact. \textbf{Right:} Random masking samples each coordinate independently as $M_i \sim \mathrm{Bernoulli}(\pi)$, corrupting features without assumptions about their content.}
    \label{fig:masking_methods}
\end{figure}

Alongside learned masking, we examine a simpler \textbf{random masking} variant, $M_{1} \sim \mathrm{Bernoulli}(\pi)$, which corrupts coordinates independently without consulting any predictor. This variant is parameter-free and serves by providing a strong task-agnostic baseline when learned masking is impractical, and it isolates the benefits of precise mask construction. Figure~\ref{fig:masking_methods} illustrates the two constructions.


\subsection{The SynIB Objective}
\label{sec:objective}

Combining the surrogate (Sec.~\ref{sec:variational}) with the masking operator (Sec.~\ref{sec:masking}) yields the SynIB objective:

\begin{tcolorbox}[colback=blue!5, colframe=blue!5, boxrule=0pt, arc=3pt, left=0pt, right=0pt, top=1pt, bottom=1pt, boxsep=0pt]
\begin{equation}
\mathcal{L}
=
\underbrace{
\mathbb{E}_{p(x_1,x_2,y)}
[-\log q_\theta(y\mid x_1,x_2)]
}_{\text{accurate prediction with intact modalities}}
+\lambda\,
\underbrace{
\mathbb{E}_{p(x_1,x_2)}
\mathbb{E}_{p(\tilde{x}_1\mid x_1)}
D_{\mathrm{KL}}\!\left(
q_\theta(\cdot\mid \tilde{x}_1,x_2)\,\|\,r(\cdot)
\right)
}_{\text{uncertainty under modality corruption}}
\label{eq:synib-loss}
\end{equation}
\end{tcolorbox}

\noindent where $q_\theta(y\mid x_1, x_2)$ is the model's predictive distribution, $r(y)$ a fixed reference distribution, and $\lambda \in \mathbb{R}^+$ the regularization strength. Figure~\ref{fig:method} illustrates the resulting training procedure: at each step, the fusion model runs forward passes on the intact pair and on each counterfactual, with the cross-entropy term applied to the first and the KL penalty to the others. 
\section{Experiments}
We evaluate SynIB on five real-world multimodal benchmarks (Sec.~\ref{sec:multibench}), comprising three MultiBench affective tasks, Hateful Memes with large pretrained backbones, and a controllable irony extension of CREMA-D we introduce, and on synthetic XOR tasks (Sec.~\ref{sec:synthetic}). We ask whether SynIB improves performance on examples requiring cross-modal reasoning, and whether targeting synergy harms accuracy on the full test set where unimodal cues already suffice.

\subsection{Real-World Multimodal Benchmarks}
\label{sec:multibench}

We evaluate on five real-world multimodal tasks. Three come from MultiBench~\citep{liang2021multibench}: UR-Funny~\citep{hasan2019urfunny} (humor detection), MUStARD~\citep{castro2019MUStARD} (sarcasm detection), and CMU-MOSI~\citep{zadeh2016mosi} (sentiment analysis). The fourth is Hateful Memes~\citep{kiela2020hateful}, where hatefulness frequently emerges only from the joint interpretation of image and text. The fifth is CREMA-D~\citep{cao2014crema}, a six-class audio--visual emotion recognition benchmark whose largely congruent signals admit strong unimodal performance. Because CREMA-D does not stress cross-modal integration, we further construct CREMA-D-Irony: for a fraction $\alpha$ of examples we replace the audio with a donor clip carrying a contradicting emotion and relabel the sample as \emph{ironic}. The rate $\alpha$ controls synergy density (App.~\ref{app:irony_dataset}).

\paragraph{Baselines.} We compare SynIB against unimodal models, late ensembling, vanilla fusion, and four modality-balancing methods: (1) D\&R~\citep{wei2024diagnosing}, which periodically re-initializes encoders; (2) MMPareto~\citep{MMPareto}, which balances unimodal and multimodal gradients via a Pareto rule; (3) ReconBoost~\citep{ReconBoost}, which alternates unimodal updates with KL reconciliation; and (4) MCR~\citep{MCR}, which adds a mutual-information regularizer to the fusion loss.
All methods share the same backbone and differ only in the training objective (full descriptions in App.~\ref{app:baselines}). For Hateful Memes we use CLIP ViT-B/16~\citep{radford2021learning} (86M) and DeBERTa-v3-base~\citep{he2021debertav3} (184M) encoders feeding a fusion Transformer.

\begin{tcolorbox}[colback=darkgray!12, colframe=darkgray!12, boxrule=0pt,
arc=3pt, left=4pt, right=4pt, top=4pt, bottom=4pt, boxsep=0pt]
 \textbf{Isolating synergy-dependent examples.} Frequently multimodal datasets test examples are solvable from a single modality. To isolate cases requiring multimodal integration, we define a \textbf{synergy subset} of test samples misclassified by every unimodal model. The subset size indicates each task's reliance on cross-modal interaction: Hateful Memes (26.5\%), MUStARD (12.7\%), UR-Funny (11.6\%), and MOSI (2.5\%). With only 17 examples, the MOSI synergy subset is too small for firm conclusions; we treat MOSI as a check that synergy-targeted training does not harm tasks where unimodal cues already suffice. CREMA-D-Irony complements this post-hoc view: the irony class is \emph{constructed} to require cross-modal integration, with $\alpha$ directly controlling its
density.
\end{tcolorbox}

\begin{figure}[t]
\centering
\pgfplotstableread[row sep=\\, col sep=&]{
alpha &
ensemble_i & ensemble_t &
vfusion_i  & vfusion_t  &
dr_i       & dr_t       &
mmp_i      & mmp_t      &
recon_i    & recon_t    &
mcr_i      & mcr_t      &
synib_i    & synib_t    &
uniV_t     & uniA_t \\
0.1 &  3.0 & 59.7 &  6.4 & 59.3 &  0.0 & 55.1 &  3.7 & 58.9 &  0.0 & 57.6 & 14.1 & 63.7 & 16.3 & 60.4 & 41.0 & 49.8 \\
0.5 & 15.0 & 58.4 & 18.3 & 57.4 &  9.0 & 55.7 & 13.0 & 58.6 & 11.6 & 57.0 & 17.6 & 60.8 & 21.7 & 59.4 & 39.9 & 46.9 \\
1.0 & 36.8 & 59.6 & 39.7 & 58.2 & 25.8 & 54.3 & 20.0 & 55.5 & 19.3 & 55.2 & 31.6 & 59.6 & 35.3 & 57.4 & 38.3 & 45.7 \\
2.0 & 54.3 & 38.3 & 50.0 & 54.5 & 45.2 & 47.0 & 41.5 & 50.2 & 45.6 & 48.8 & 53.5 & 55.5 & 53.0 & 55.3 & 28.0 & 39.6 \\
}{\datatable}

\usepgfplotslibrary{groupplots}

\pgfplotsset{
  mypanel/.style={
    width=0.48\textwidth,
    height=0.38\textwidth,
    xmode=log, log basis x=10,
    xmin=0.085, xmax=2.6,
    xtick={0.1,0.5,1,2},
    xticklabels={0.1, 0.5, 1, 2},
    xlabel={Irony rate $\alpha$},
    grid=major,
    major grid style={draw=gray!20},
    tick align=inside,
    xtick pos=left,
    ytick pos=left,
    tick label style={font=\scriptsize, inner sep=1pt},
    label style={font=\small},
    title style={font=\small\bfseries, at={(0.02,0.97)}, anchor=north west, text=gray!60!black},
    xlabel style={font=\small, yshift=2pt},
    ylabel style={font=\small, yshift=-2pt},
    enlarge x limits=0.01,
    clip mode=individual,
  }
}

\begin{tikzpicture}

\begin{axis}[
  hide axis,
  scale only axis,
  height=0pt, width=0pt,
  xmin=0, xmax=1, ymin=0, ymax=1,
  legend to name=mylegend,
  legend columns=1,
  legend style={
    font=\scriptsize,
    draw=none,
    fill=none,
    inner sep=4pt,
    row sep=0.4ex,
    legend cell align=left,
  },
]
\addlegendimage{gray, densely dashed, line width=0.8pt}                                        \addlegendentry{Uni-V}
\addlegendimage{gray, densely dotted, line width=0.8pt}                                        \addlegendentry{Uni-A}
\addlegendimage{blue,           mark=triangle*, semithick, mark size=2}                        \addlegendentry{Ensemble}
\addlegendimage{orange,         mark=square*,   semithick, mark size=2}                        \addlegendentry{Vanilla fusion}
\addlegendimage{green!55!black, mark=diamond*,  semithick, mark size=2}                        \addlegendentry{D\&R}
\addlegendimage{red!65!black,   mark=*,         semithick, mark size=2}                        \addlegendentry{MMPareto}
\addlegendimage{purple,         mark=o,         semithick, mark size=2}                        \addlegendentry{ReconBoost}
\addlegendimage{teal!70!black,  mark=star,      semithick, mark size=2.5}                      \addlegendentry{MCR}
\addlegendimage{black, solid,   mark=*, very thick, mark size=2.5, mark options={fill=orange}} \addlegendentry{\textbf{SynIB}}
\end{axis}

\begin{groupplot}[
  group style={
    group size=2 by 1,
    horizontal sep=0.9cm,
  },
  mypanel,
]

\nextgroupplot[
  title={Irony recognition},
  ylabel={Irony-class F1},
  ymin=-2.0, ymax=62,
  xmin=0, xmax=2,
  ytick={0,10,20,30,40,50,60},
  yticklabels={0,10,20,30,40,50,60},
]
\addplot+[blue,           mark=triangle*, semithick,  mark size=2  ] table[x=alpha,y=ensemble_i]{\datatable};
\addplot+[orange,         mark=square*,   semithick,  mark size=2  ] table[x=alpha,y=vfusion_i] {\datatable};
\addplot+[green!55!black, mark=diamond*,  semithick,  mark size=2  ] table[x=alpha,y=dr_i]      {\datatable};
\addplot+[red!65!black,   mark=*,         semithick,  mark size=2  ] table[x=alpha,y=mmp_i]     {\datatable};
\addplot+[purple,         mark=o,         semithick,  mark size=2  ] table[x=alpha,y=recon_i]   {\datatable};

\nextgroupplot[
  title={Overall performance},
  ylabel={Total F1},
  ymin=34, ymax=67,
  xmin=0, xmax=2,
  ytick={35,40,45,50,55,60,65},
  yticklabels={35,40,45,50,55,60,65},
]
\addplot+[blue,           mark=triangle*, semithick,  mark size=2  ] table[x=alpha,y=ensemble_t]{\datatable};
\addplot+[orange,         mark=square*,   semithick,  mark size=2  ] table[x=alpha,y=vfusion_t] {\datatable};
\addplot+[green!55!black, mark=diamond*,  semithick,  mark size=2  ] table[x=alpha,y=dr_t]      {\datatable};
\addplot+[red!65!black,   mark=*,         semithick,  mark size=2  ] table[x=alpha,y=mmp_t]     {\datatable};
\addplot+[purple,         mark=o,         semithick,  mark size=2  ] table[x=alpha,y=recon_t]   {\datatable};
\addplot+[teal!70!black,  mark=star,      semithick,  mark size=2.5] table[x=alpha,y=mcr_t]     {\datatable};
\addplot+[black, solid,   mark=*,         very thick, mark size=2.5, mark options={fill=orange}] table[x=alpha,y=synib_t]{\datatable};
\addplot+[gray, densely dashed, no marks, line width=0.8pt] table[x=alpha,y=uniV_t]{\datatable};
\addplot+[gray, densely dotted, no marks, line width=0.8pt] table[x=alpha,y=uniA_t]{\datatable};

\end{groupplot}

\node[anchor=west, inner sep=0pt]
  at ($(group c2r1.east) + (0.1cm, 0)$)
  {\pgfplotslegendfromname{mylegend}};

\end{tikzpicture}

\caption{%
  F1 scores on the CREMA-D irony recognition task under varying irony rates $\alpha$.
  \textbf{Left:} irony-class F1. \textbf{Right:} total F1.
  Standard fusion and prior balancing methods yield unstable gains when irony is rare, whereas \textbf{SynIB} consistently improves irony detection across all regimes with only minor trade-offs in overall performance, often achieving the best total F1.
}

\label{fig:crema_irony_rates}
\end{figure}
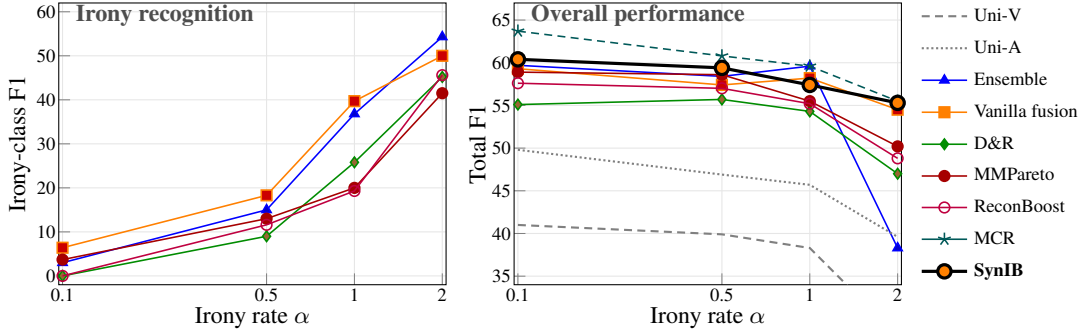

\paragraph{Results on CREMA-D-Irony.}
Figure~\ref{fig:crema_irony_rates} reports irony-class F1 (left) and
total F1 (right) across $\alpha$. SynIB outperforms MCR, the strongest baseline, on irony F1 at every $\alpha$, most clearly at $\alpha{=}0.5$ (21.7 vs.\ 17.6) and with consistent margins at $\alpha{=}0.1$ (16.3 vs.\ 14.1) and $\alpha{=}1.0$ (35.3 vs.\ 31.6). Vanilla fusion and ensembling become competitive only when synergy is abundant ($\alpha{\ge}1$). SynIB stays within 1.0--3.3 points of the best total F1 at every $\alpha$ (full numbers in App.~\ref{app:crema_irony_full}).


\paragraph{Results on MultiBench and Hateful Memes.}
Figure~\ref{fig:synib_bar_dual} reports accuracy on the synergy subset and on the full test set. On the synergy subset, SynIB achieves the largest gains across all four benchmarks, with learned masking improving over the strongest baseline by $+3.0$ on Hateful Memes, $+7.8$ on MUStARD, and $+3.6$ on UR-Funny; random masking is strong on Hateful Memes ($+3.0$) but more variable elsewhere, suggesting that the targeted mask is what reliably drives improvements on synergy-required examples. Modality-balancing methods (D\&R, MMPareto, ReconBoost, MCR) yield inconsistent gains on the synergy subset, consistent with our argument that rebalancing alone does not induce synergistic learning. On the full test set, SynIB remains the top performer on Hateful Memes ($+1.1$ random, $+0.3$ learned) and MUStARD ($+3.8$ learned), and stays within  $1$ point on UR-Funny ($-0.9$) and MOSI ($+0.6$), indicating no cost to overall accuracy.

\begin{figure}[t]
\centering

\definecolor{EnsGray}    {RGB}{205,205,205}
\definecolor{VanSlate}   {RGB}{178,178,178}
\definecolor{DRSage}     {RGB}{150,150,150}
\definecolor{ParetoTeal} {RGB}{122,122,122}
\definecolor{ReconMauve} {RGB}{ 92, 92, 92}
\definecolor{MCRViolet}  {RGB}{ 58, 58, 58}
\definecolor{SynRandBlue} {RGB}{ 88,158,218}
\definecolor{SynLearnNavy}{RGB}{ 18, 68,145}
\definecolor{GridGray}{RGB}{225,225,225}
\definecolor{AxisGray}{RGB}{118,118,118}

\resizebox{\linewidth}{!}{%
\begin{tikzpicture}[font=\small]

\begin{axis}[
    name=leftax,
    at={(-1.3cm,2.55cm)}, anchor=center,
    width=8.4cm, height=4.95cm,
    ybar, bar width=3.6pt,
    ymin=0, ymax=68,
    ytick={0,10,20,30,40,50,60},
    yticklabel style={font=\scriptsize},
    title={\scriptsize\bfseries(a) Synergy Subset Accuracy},
    title style={yshift=-2pt},
    symbolic x coords={UR-Funny,MUStARD,MOSI,HatefulMemes},
    xtick=data, xticklabel style={font=\scriptsize},
    enlarge x limits=0.10,
    axis x line*=bottom, axis y line*=left,
    axis line style={draw=AxisGray, line width=0.5pt},
    tick style={draw=AxisGray},
    ymajorgrids=true, xmajorgrids=false,
    grid style={draw=GridGray, line width=0.35pt},
    clip=false,
]
\addplot[draw=none,fill=EnsGray]    coordinates {(UR-Funny,0.9) (MUStARD,45.1)(MOSI,0.0) (HatefulMemes,29.6)};
\addplot[draw=none,fill=VanSlate]   coordinates {(UR-Funny,19.2)(MUStARD,25.4)(MOSI,31.4)(HatefulMemes,33.8)};
\addplot[draw=none,fill=DRSage]     coordinates {(UR-Funny,15.2)(MUStARD,16.5)(MOSI,24.1)(HatefulMemes,42.1)};
\addplot[draw=none,fill=ParetoTeal] coordinates {(UR-Funny,15.9)(MUStARD,40.1)(MOSI,25.1)(HatefulMemes,42.4)};
\addplot[draw=none,fill=ReconMauve] coordinates {(UR-Funny,12.7)(MUStARD,17.2)(MOSI,30.2)(HatefulMemes,42.3)};
\addplot[draw=none,fill=MCRViolet]  coordinates {(UR-Funny,11.0)(MUStARD,39.6)(MOSI,29.0)(HatefulMemes,45.9)};

\addplot[
    draw=none, fill=SynRandBlue,
    nodes near coords,
    nodes near coords align=vertical,
    every node near coord/.append style={
        font=\tiny\bfseries, rotate=90, anchor=west,
        color=SynRandBlue!80!black, inner sep=1.2pt
    },
    point meta=explicit symbolic,
] coordinates {
    (UR-Funny,15.4)     [$-3.8$]
    (MUStARD,47.5)      [$+2.4$]
    (MOSI,35.3)         [$+3.9$]
    (HatefulMemes,48.9) [$+3.0$]
};

\addplot[
    draw=none, fill=SynLearnNavy,
    nodes near coords,
    nodes near coords align=vertical,
    every node near coord/.append style={
        font=\tiny\bfseries, rotate=90, anchor=west,
        color=SynLearnNavy, inner sep=1.2pt
    },
    point meta=explicit symbolic,
] coordinates {
    (UR-Funny,22.8)     [$+3.6$]
    (MUStARD,52.9)      [$+7.8$]
    (MOSI,37.2)         [$+5.8$]
    (HatefulMemes,47.0) [$+1.1$]
};

\node[anchor=south east, font=\scriptsize, inner sep=1pt]
    at (rel axis cs:0.0, 1.0) {Acc (\%)};
\end{axis}

\begin{axis}[
    name=rightax,
    at={(6.55cm,2.55cm)}, anchor=center,
    width=8.4cm, height=4.95cm,
    ybar, bar width=3.6pt,
    ymin=48, ymax=78,
    ytick={50,55,60,65,70,75},
    yticklabel style={font=\scriptsize},
    title={\scriptsize\bfseries(b) Whole-test Accuracy},
    title style={yshift=-2pt},
    symbolic x coords={UR-Funny,MUStARD,MOSI,HatefulMemes},
    xtick=data, xticklabel style={font=\scriptsize},
    enlarge x limits=0.18,
    axis x line*=bottom, axis y line*=left,
    axis line style={draw=AxisGray, line width=0.5pt},
    tick style={draw=AxisGray},
    ymajorgrids=true, xmajorgrids=false,
    grid style={draw=GridGray, line width=0.35pt},
    clip=false,
]
\addplot[draw=none,fill=EnsGray]    coordinates {(UR-Funny,61.3)(MUStARD,58.0)(MOSI,72.8)(HatefulMemes,55.2)};
\addplot[draw=none,fill=VanSlate]   coordinates {(UR-Funny,62.3)(MUStARD,57.5)(MOSI,72.8)(HatefulMemes,66.4)};
\addplot[draw=none,fill=DRSage]     coordinates {(UR-Funny,63.6)(MUStARD,54.5)(MOSI,73.0)(HatefulMemes,67.2)};
\addplot[draw=none,fill=ParetoTeal] coordinates {(UR-Funny,62.8)(MUStARD,58.8)(MOSI,72.6)(HatefulMemes,68.0)};
\addplot[draw=none,fill=ReconMauve] coordinates {(UR-Funny,63.2)(MUStARD,57.5)(MOSI,74.7)(HatefulMemes,66.9)};
\addplot[draw=none,fill=MCRViolet]  coordinates {(UR-Funny,62.4)(MUStARD,60.3)(MOSI,73.7)(HatefulMemes,68.7)};

\addplot[
    draw=none, fill=SynRandBlue,
    nodes near coords,
    nodes near coords align=vertical,
    every node near coord/.append style={
        font=\tiny\bfseries, rotate=90, anchor=west,
        color=SynRandBlue!80!black, inner sep=1.2pt
    },
    point meta=explicit symbolic,
] coordinates {
    (UR-Funny,63.6)     [$=$]
    (MUStARD,62.2)      [$+1.9$]
    (MOSI,75.0)         [$+0.3$]
    (HatefulMemes,69.8) [$+1.1$]
};

\addplot[
    draw=none, fill=SynLearnNavy,
    nodes near coords,
    nodes near coords align=vertical,
    every node near coord/.append style={
        font=\tiny\bfseries, rotate=90, anchor=west,
        color=SynLearnNavy, inner sep=1.2pt
    },
    point meta=explicit symbolic,
] coordinates {
    (UR-Funny,62.7)     [$-0.9$]
    (MUStARD,64.1)      [$+3.8$]
    (MOSI,75.3)         [$+0.6$]
    (HatefulMemes,69.0) [$+0.3$]
};

\node[anchor=south east, font=\scriptsize, inner sep=1pt]
    at (rel axis cs:0.0, 1.0) {Acc (\%)};
\end{axis}

\begin{scope}[shift={(-5.0,-0.12)}]

\fill[EnsGray]   (0.00,-0.07) rectangle +(0.28,0.17);
\draw[EnsGray!55!black,   line width=0.22pt] (0.00,-0.07) rectangle +(0.28,0.17);
\node[anchor=west,font=\scriptsize] at (0.37,0.02) {Ensemble};

\fill[VanSlate]  (3.10,-0.07) rectangle +(0.28,0.17);
\draw[VanSlate!55!black,  line width=0.22pt] (3.10,-0.07) rectangle +(0.28,0.17);
\node[anchor=west,font=\scriptsize] at (3.47,0.02) {Vanilla Fusion};

\fill[DRSage]    (7.20,-0.07) rectangle +(0.28,0.17);
\draw[DRSage!55!black,    line width=0.22pt] (7.20,-0.07) rectangle +(0.28,0.17);
\node[anchor=west,font=\scriptsize] at (7.57,0.02) {D\&R};

\fill[ParetoTeal](11.30,-0.07) rectangle +(0.28,0.17);
\draw[ParetoTeal!55!black,line width=0.22pt] (11.30,-0.07) rectangle +(0.28,0.17);
\node[anchor=west,font=\scriptsize] at (11.67,0.02) {MMPareto};

\fill[ReconMauve](0.00,-0.44) rectangle +(0.28,0.17);
\draw[ReconMauve!55!black,line width=0.22pt] (0.00,-0.44) rectangle +(0.28,0.17);
\node[anchor=west,font=\scriptsize] at (0.37,-0.35) {ReconBoost};

\fill[MCRViolet] (3.10,-0.44) rectangle +(0.28,0.17);
\draw[MCRViolet!55!black, line width=0.22pt] (3.10,-0.44) rectangle +(0.28,0.17);
\node[anchor=west,font=\scriptsize] at (3.47,-0.35) {MCR};

\fill[SynRandBlue](7.20,-0.44) rectangle +(0.28,0.17);
\draw[SynRandBlue!55!black,line width=0.22pt] (7.20,-0.44) rectangle +(0.28,0.17);
\node[anchor=west,font=\scriptsize] at (7.57,-0.35) {$\mathrm{SynIB}\ M_{\mathrm{Random}}$};

\fill[SynLearnNavy](11.30,-0.44) rectangle +(0.28,0.17);
\draw[SynLearnNavy!55!black,line width=0.22pt] (11.30,-0.44) rectangle +(0.28,0.17);
\node[anchor=west,font=\scriptsize] at (11.67,-0.35) {$\mathrm{SynIB}\ M_{\mathrm{Learned}}$};

\end{scope}

\path[use as bounding box] (0.0,-1.15) rectangle (0.5,5.35);

\end{tikzpicture}%
}
\caption{Comparison across real-world multimodal benchmarks.
\textbf{(a)} Synergy-subset accuracy on UR-Funny, MUStARD, MOSI, and
Hateful Memes (samples misclassified by every unimodal model).
\textbf{(b)} Whole-test accuracy. Baselines comprise late ensembling,
vanilla fusion, and four modality-balancing methods:
D\&R~\citep{wei2024diagnosing}, MMPareto~\citep{MMPareto},
ReconBoost~\citep{ReconBoost}, and MCR~\citep{MCR}; SynIB with
($M_{\mathrm{Random}}$) and ($M_{\mathrm{Learned}}$) masking.
Annotations above SynIB bars give the difference relative to the best
baseline. SynIB consistently improves synergy-subset accuracy on all
four while remaining competitive on the full test set.}
\label{fig:synib_bar_dual}
\end{figure}

Across five benchmarks spanning humor, sarcasm, sentiment, hate speech, and emotion-irony, SynIB consistently improves performance on the examples that demonstrably require cross-modal reasoning, while remaining competitive, and often best, on the full test distribution. The pattern is most pronounced on tasks with substantial synergy content (CREMA-D-Irony, MUStARD, Hateful Memes) and naturally smaller on tasks where unimodal shortcuts already solve most of the problem.


\subsection{Synthetic Experiments}
\label{sec:synthetic}

We design two bimodal XOR tasks. \emph{Spurious XOR} probes
resistance to unimodal shortcuts: the label is the XOR of two
modality-specific bits (neither modality alone informative), but one modality additionally carries a label-correlated feature available only at training, letting the model fit training perfectly without the cross-modal signal. \emph{PID-Controlled XOR} sweeps the density of each PID source (unique-to-$X_1$, redundant, synergistic). Full setup in App.~\ref{app:synthetic_details}.

\begin{wrapfigure}[16]{r}{0.5\linewidth}
  \vspace{-6mm}
  \centering
  \includegraphics[width=\linewidth]{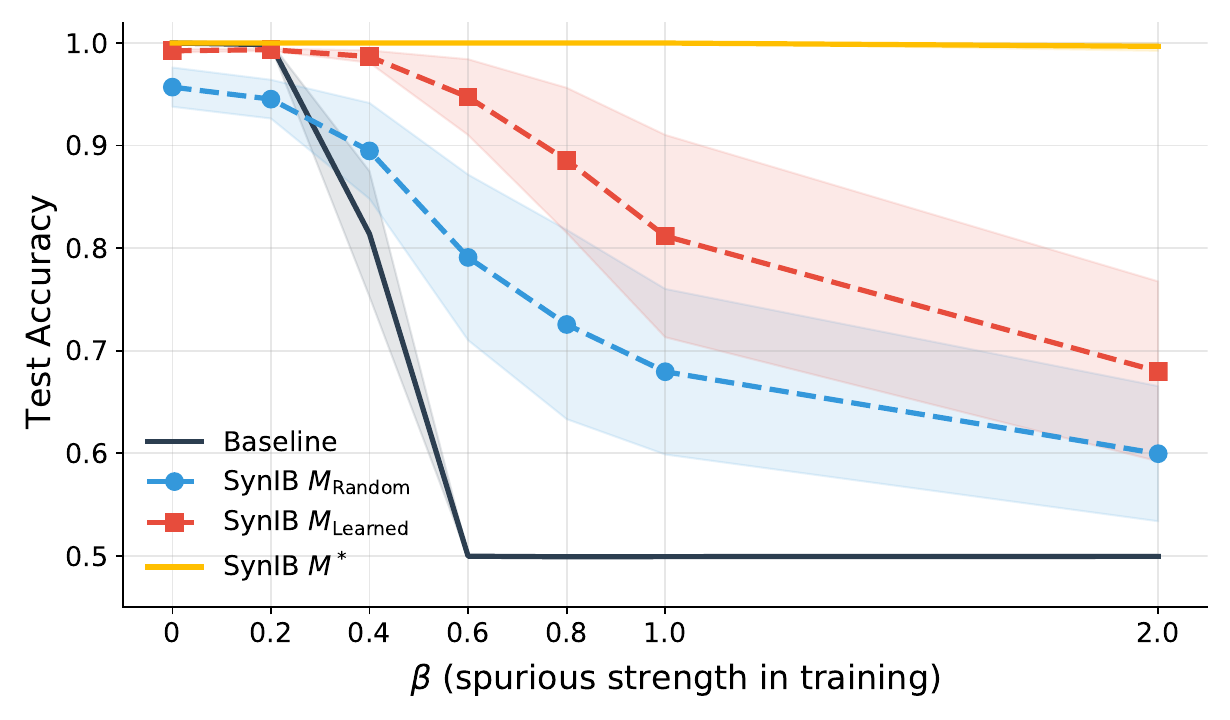}
  \vspace{-14pt}
  \caption{Test accuracy vs.\ spurious correlation strength $\beta$ on bimodal XOR. Vanilla fusion collapses to chance as the shortcut strengthens. SynIB with oracle masking stays at $\sim$100\%, while learned and random masks degrade gracefully, with the learned mask consistently closer to the oracle.}
  \vspace{10mm}
  \label{fig:spurious_correlations}
\end{wrapfigure}

\paragraph{Robustness to spurious shortcuts.} One modality contains a spurious feature linearly correlated with the label at training strength $\beta$ but uncorrelated at test. As $\beta$ grows, the shortcut becomes increasingly attractive during training but provides no test-time signal. Figure~\ref{fig:spurious_correlations} shows that vanilla fusion collapses to chance once $\beta \geq 0.6$, locking onto the shortcut and discarding the cross-modal signal. SynIB with oracle masking maintains $\sim$100\% across all $\beta$, confirming that isolating synergistic information suffices. Without oracle annotations, both random and learned masks substantially improve robustness at every $\beta$, with the learned mask consistently closer to the oracle by identifying the features the unimodal head exploits and corrupting their complement.

\begin{figure}[t]
    \centering
    \includegraphics[width=0.99\linewidth]{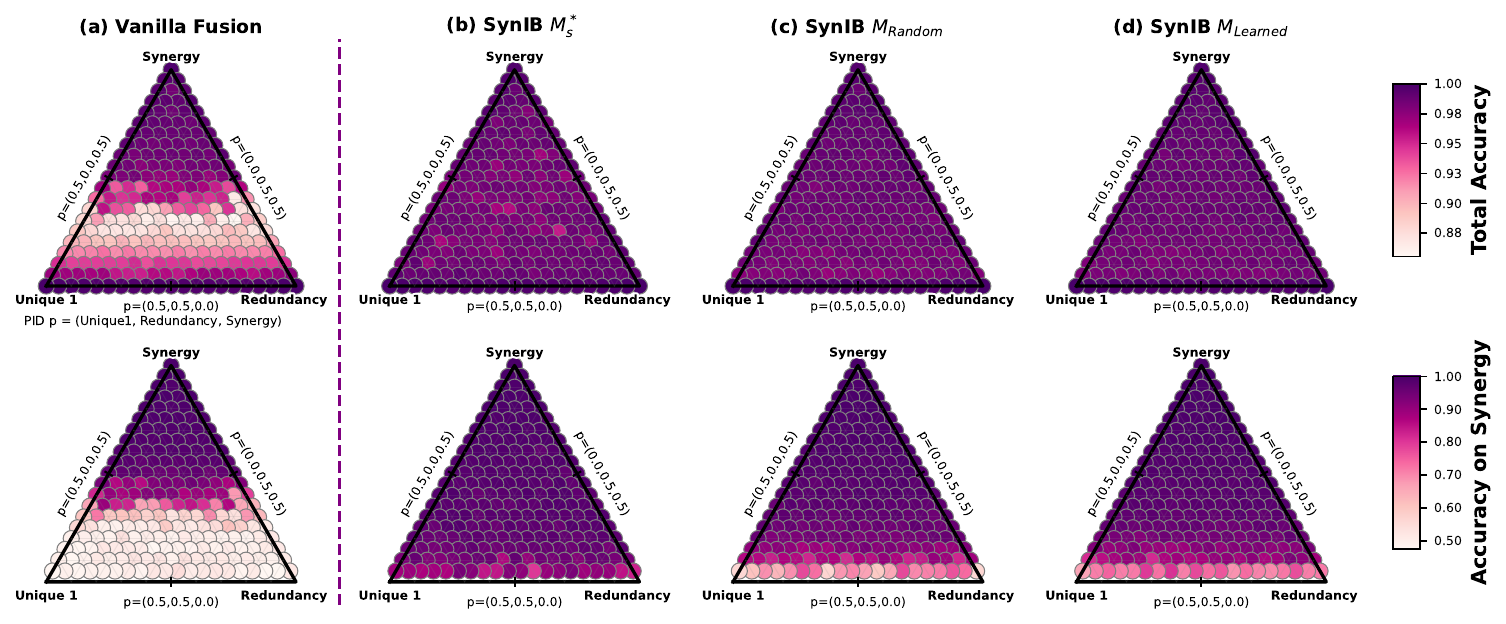}
    \caption{
    \textbf{Performance across PID-controlled data regimes on the synthetic XOR task.}
    Each triangle is a probability simplex over information types $(p_{\mathrm{U1}}, p_{\mathrm{Red}}, p_{\mathrm{Syn}})$; top row shows total accuracy, bottom row accuracy on synergistic examples only.
    Vanilla fusion (a) degrades sharply as synergy dominates. SynIB with oracle masking (b) resolves the task uniformly. Random masking (c) provides inconsistent gains, while learned masking (d) closely recovers oracle performance.
    }
    \label{fig:pid_xor}
\end{figure}

\paragraph{Coverage across PID compositions.} Each example is generated from exactly one active source (unique, redundant, or synergistic), enabling controlled interpolation between unimodal-dominated and synergy-dominated regimes. Figure~\ref{fig:pid_xor} reports performance across the information simplex. Without SynIB (Fig.~\ref{fig:pid_xor}a), accuracy degrades sharply as synergistic structure becomes dominant; with oracle masking (\ref{fig:pid_xor}b) it remains uniformly high. Random masking (\ref{fig:pid_xor}c) yields inconsistent gains, while the learned mask (\ref{fig:pid_xor}d) closely matches the oracle even when synergy is scarce, indicating that SynIB recovers synergistic structure from data without requiring annotations.

\begin{figure}[t]
    \centering
    \includegraphics[width=\linewidth]{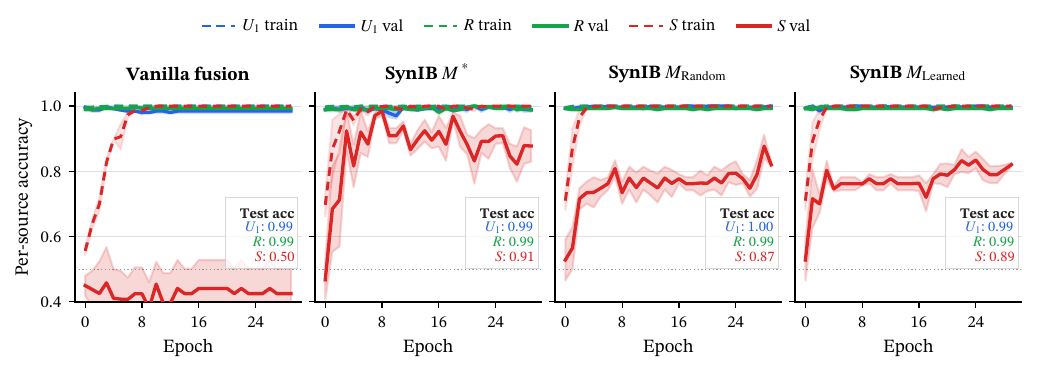}
\caption{\textbf{PID-XOR training dynamics across all four methods.} Per-source training (solid) and validation (dashed) accuracy across 30 epochs, mean $\pm$ standard error over three seeds. Sources are colored by type: unique-to-modality-1 ($U_1$, blue), redundant ($R$, green), and synergistic ($S$, red). Final synergy test accuracies (annotated per panel): $0.50$ vanilla, $0.91$ oracle, $0.87$ random, $0.89$ learned. PID mixture $(p_{U_1}, p_{U_2}, p_R, p_S) = (0.45, 0, 0.45, 0.10)$. Vanilla fusion (left) fits $U_1$ and $R$ on both splits, but synergy validation stalls at chance while training accuracy climbs, the signature of memorization on the scarce synergistic examples. The three SynIB variants ($M^\star$, $M_{\mathrm{Random}}$, $M_{\mathrm{Learned}}$) close this gap, lifting synergy validation accuracy without degrading $U_1$ or $R$.}
  \label{fig:pid-xor-training-dynamics}     
\end{figure}

\paragraph{Per-source training dynamics.}
Figure~\ref{fig:pid-xor-training-dynamics} disaggregates accuracy by PID source across training, directly testing the diagnosis from Sec.~\ref{sec:failure_mode}: that synergy fails through overfitting on a scarce subset rather than through gradient interference or capacity limits. Vanilla fusion fits $U_1$ and $R$ on both splits, but synergy shows the signature train--validation divergence, with training accuracy climbing while validation stalls at chance. The three SynIB variants close this gap: synergy train and validation track each other throughout training, with learned masking ($0.89$) tracking the oracle ($0.91$) within two points and random masking ($0.87$) close behind.

\section{Conclusion}
Multimodal models often default to unimodal shortcuts during training,
even when the task rewards combining modalities. We argued this is an
optimization problem: synergistic cues are scarce, so the model fits
dominant unimodal signals first and never learns cross-modal
interaction. SynIB addresses this by penalizing the model when its
predictions remain confident under modality corruption, pushing it
toward signals that emerge only from joint use of both modalities.
Across synthetic tasks, and five
real-world datasets including the controllable irony construction of CREMA-D and the Hateful Memes with pretrained backbones,
SynIB consistently improves accuracy on examples requiring cross-modal
reasoning while remaining competitive on the full test set. See
App.~\ref{app:limitations} for limitations.

\paragraph{Broader Impact.} Synergistic information underlies many of the settings where multimodal learning is most consequential, and methods that reliably train models to use cross-modal interactions, rather than defaulting to whichever modality is easiest to fit, expand the set of problems where multimodal learning can deliver on its premise. Better synergy does not, on its own, make a model trustworthy: a model that integrates modalities more effectively can still inherit biases from its training data, fail under distribution shift, or be applied in settings where higher accuracy is not the primary concern. We see synergy-targeted training as one ingredient in responsible multimodal learning, useful where cross-modal reasoning is the bottleneck and best paired with the auditing practices appropriate to each application.

\clearpage
{\small
\bibliographystyle{plainnat}
\bibliography{citations_MCR}  
}

\clearpage
\newpage
\appendix
\vspace{2.5em}
{\Large\bfseries \centering Supplementary Material\par}
\vspace{2.5em}

\section{Limitations}
\label{app:limitations}
We discuss four limitations of SynIB that we believe are important for interpreting our results and identifying directions for future work. Limitations of the NTK-based motivating analysis in Sec.~\ref{sec:failure_mode} are discussed separately in App.~\ref{app:ntk_pid}.

\paragraph{Dependence on the latent representation} SynIB operates on representations produced by modality-specific encoders, and its effectiveness ultimately depends on whether synergistic features are expressible in the latent space those encoders produce. When the encoders strip out cross-modal structure during their forward pass, no loss-level objective can recover it. Self-supervised pretrained encoders such as CLIP partially mitigate this by preserving richer per-modality features, and our Hateful Memes experiments suggest that SynIB transfers cleanly to this setting. A more systematic study of how pretraining choices affect the synergy a fusion model can ultimately learn remains open.

\paragraph{Sensitivity of the learned mask} The adversarial mask $M_\psi$ is trained jointly with the SynIB objective and depends on hyperparameters governing sparsity ($\lambda_M$), optimization dynamics, and the unimodal predictor used to define the adversary. In our experiments these settings transferred reasonably across datasets without per-task tuning, but in regimes with very small synergy subsets, mask training can become unstable and may collapse to either trivial (no features kept) or near-identity (all features kept) solutions. Random masking, while less precise, is more robust in these cases and provides a useful fallback.

\paragraph{Variational surrogate rather than exact decomposition} Our objective is motivated by partial information decomposition, but we optimize a tractable variational surrogate of the Mutual Information Perturbed Difference rather than computing PID quantities directly. The surrogate coincides with MIPD when the variational family contains the true predictive distributions, and tracks the same direction in the loss landscape more generally; in particular, it is not guaranteed to be a one-sided bound on MIPD. SynIB therefore pushes the model toward predictions that depend on cross-modal interaction, but does not produce a formal decomposition of $I(Y; X_1, X_2)$ into unique, redundant, and synergistic components. The surrogate is sufficient for shaping training but does not, on its own, yield an explicit attribution of the learned model's predictions to unique, redundant, and synergistic components; that attribution is a separate analytical question we leave to future work.

\paragraph{Training-time overhead} Each SynIB training step requires forward passes through the fusion model on both the intact inputs and one or more counterfactual inputs (one per masked modality), increasing per-step compute by roughly $1{+}n$ where $n$ is the number of modalities. The learned-masking variant additionally trains the mask network jointly. In our experiments this overhead was tolerable, but for very large backbones or high-modality settings, more efficient approximations, for example sampling a subset of modalities to mask per batch, would be valuable. We leave this to future work.
\subsection{NTK Analysis with SynIB}
\label{app:ntk_synib}

Section~\ref{sec:failure_mode} used the empirical NTK to diagnose
why synergistic examples fail to generalize under vanilla training: the
largest per-source learning signal $\lambda_g$ is allocated to synergy
and no destructive interference appears between PID sources, yet
synergistic validation loss rises while training loss drops. This
appendix repeats the same diagnostics on a model trained with SynIB,
verifying that the objective shifts these quantities in the direction
the motivating analysis prescribes.

\paragraph{Setup.} We train on the PID-Controlled XOR data
(App.~\ref{app:pid_xor}) at composition
$(p_{U_1}, p_R, p_S) = (0.45, 0.45, 0.10)$, identical to the vanilla run
in Fig.~\ref{fig:pid_ntk_geometry}, and add SynIB with the learned-mask inner
loop ($\lambda = 10$). Architecture, optimiser, batching, and seeds are
unchanged.

\begin{figure}
    \centering
    \includegraphics[width=1\linewidth]{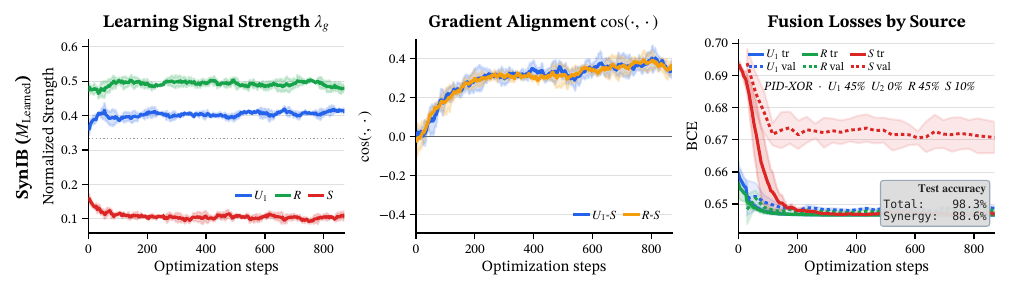}
    \caption{\textbf{Gradient geometry across PID sources under SynIB.}
    Training augments the vanilla setup with the SynIB learned-mask
    inner loop ($\lambda{=}10$); data, architecture, and optimiser are
    otherwise identical to Fig.~\ref{fig:pid_ntk_geometry}.
    \textbf{Left:} per-source learning signal strength $\lambda_g$ over
    training. SynIB redistributes mass toward $U_1$ and $R$, leaving
    synergy with the smallest share without starving it.
    \textbf{Center:} gradient alignment between source pairs stays near
    zero throughout training, ruling out destructive competition.
    \textbf{Right:} fusion BCE by source on intact inputs (solid train,
    dashed validation). Train and validation curves track closely on
    every source; the synergistic train--validation gap that opened
    under vanilla is now closed. Test accuracy: synergy $88.6\%$, total
    $98.3\%$.}
    \label{fig:pid_ntk_synib}
\end{figure}

\paragraph{Findings.} Three changes appear in
Fig.~\ref{fig:pid_ntk_synib}, each consistent with the SynIB design.
First, the per-source learning signal redistributes: $\lambda_{U_1}$ and
$\lambda_R$ rise while $\lambda_S$ falls (left), so SynIB de-emphasizes
the scarce synergistic examples that were being memorized under vanilla
without starving any source. Second, gradient alignment between sources
remains near zero throughout training (center), confirming that the
redistribution does not introduce destructive competition. Third, and
most importantly, the train--validation gap on the synergistic source
closes (right): BCE on synergistic train and validation now track each
other, in contrast to the diverging curves under vanilla. Combined with
test accuracy of $88.6\%$ on the synergy slice ($98.3\%$ overall), these
findings indicate that SynIB reshapes the per-source signal in the way
the motivating analysis prescribes, rather than simply injecting noise
into training.

\section{Variational Surrogate for MIPD}
\label{app:mipd-bounds}

\subsection{Derivation of the SynIB Surrogate}

This appendix derives the SynIB training objective as a tractable surrogate for $\mathrm{MIPD}_1 = H(Y\mid \tilde{X}_1, X_2) - H(Y\mid X_1, X_2)$. The two conditional entropies in this expression are not directly computable for high-dimensional inputs, since they require access to the true predictive distributions $p(y\mid x_1, x_2)$ and $p(y\mid \tilde{x}_1, x_2)$. We approximate each term using the variational distribution $q_\theta(y\mid \cdot)$ produced by the model, with $r(y)$ denoting a fixed reference distribution over $\mathcal{Y}$ chosen independently of the inputs and of the model parameters.

\paragraph{Upper bound on $H(Y\mid X_1, X_2)$ via cross-entropy.}
For the entropy under intact inputs, the standard variational upper bound due to \citet{barber2004algorithm} applies. For any variational distribution $q_\theta(y\mid x_1, x_2)$,
\begin{align}
\mathbb{E}_{p(x_1, x_2, y)}[-\log q_\theta(y\mid x_1, x_2)]
&= H(Y\mid X_1, X_2) + \mathbb{E}_{p(x_1, x_2)}\, D_{\mathrm{KL}}\!\left(p(\cdot\mid x_1, x_2)\,\|\,q_\theta(\cdot\mid x_1, x_2)\right),
\label{eq:ba-decomp}
\end{align}
which follows directly from expanding the cross-entropy and identifying the KL divergence between the true and variational conditionals. Since $D_{\mathrm{KL}} \ge 0$, this yields
\begin{align}
H(Y\mid X_1, X_2) \le \mathbb{E}_{p(x_1, x_2, y)}[-\log q_\theta(y\mid x_1, x_2)],
\label{eq:ba-upper}
\end{align}
with equality if and only if $q_\theta(\cdot\mid x_1, x_2) = p(\cdot\mid x_1, x_2)$ almost everywhere. The bound's tightness is governed by the variational approximation error $\mathbb{E}_{p(x_1, x_2)} D_{\mathrm{KL}}(p\,\|\,q_\theta)$, which the cross-entropy training objective itself minimizes.

\paragraph{Reference-based identity for $H(Y\mid \tilde{X}_1, X_2)$.}
For the entropy under modality corruption, we use an exact identity that re-expresses conditional entropy in terms of a divergence to the reference $r$. Starting from the definition of conditional entropy and adding and subtracting $\log r(y)$,
\begin{align}
H(Y\mid \tilde{X}_1, X_2)
&= -\mathbb{E}_{p(\tilde{x}_1, x_2, y)}[\log p(y\mid \tilde{x}_1, x_2)] \nonumber \\
&= -\mathbb{E}_{p(\tilde{x}_1, x_2, y)}\!\left[\log \frac{p(y\mid \tilde{x}_1, x_2)}{r(y)}\right] - \mathbb{E}_{p(y)}[\log r(y)] \nonumber \\
&= C_r - \mathbb{E}_{p(\tilde{x}_1, x_2)}\, D_{\mathrm{KL}}\!\left(p(\cdot\mid \tilde{x}_1, x_2)\,\|\,r(\cdot)\right),
\label{eq:ref-identity-app}
\end{align}
where $C_r := \mathbb{E}_{p(y)}[-\log r(y)]$ is the cross-entropy of the marginal $p(y)$ relative to $r$ and is constant with respect to $\theta$. This identity, used in the Information Bottleneck framework~\citep{tishby2000information,alemi2016deep}, expresses the conditional entropy as a constant minus the average divergence between the true conditional and the reference: when $p(\cdot\mid \tilde{x}_1, x_2)$ is far from $r$, the conditional entropy is small (predictions are sharp), and conversely.

\paragraph{Variational surrogate.}
Eq.~\eqref{eq:ref-identity-app} is exact but not directly usable, since the KL divergence involves the unknown $p(\cdot\mid \tilde{x}_1, x_2)$. We obtain a tractable expression by replacing the true conditional with the variational $q_\theta(\cdot\mid \tilde{x}_1, x_2)$:
\begin{align}
\widehat{H}_\theta(Y\mid \tilde{X}_1, X_2)
&:=
C_r - \mathbb{E}_{p(\tilde{x}_1, x_2)}\, D_{\mathrm{KL}}\!\left(q_\theta(\cdot\mid \tilde{x}_1, x_2)\,\|\,r(\cdot)\right).
\label{eq:H-surrogate}
\end{align}
The substitution defines a surrogate for $H(Y\mid \tilde{X}_1, X_2)$ that is computable from the model's predictions on samples drawn from $p(\tilde{x}_1, x_2)$. By construction, $\widehat{H}_\theta(Y\mid \tilde{X}_1, X_2) = H(Y\mid \tilde{X}_1, X_2)$ exactly when $q_\theta(\cdot\mid \tilde{x}_1, x_2) = p(\cdot\mid \tilde{x}_1, x_2)$.

\paragraph{Combining the two terms.}
Substituting Eq.~\eqref{eq:ba-upper} for $H(Y\mid X_1, X_2)$ and Eq.~\eqref{eq:H-surrogate} for $H(Y\mid \tilde{X}_1, X_2)$ in $\mathrm{MIPD}_1$ yields the SynIB surrogate
\begin{align}
\widehat{\mathrm{MIPD}}_1
&= \widehat{H}_\theta(Y\mid \tilde{X}_1, X_2) - \mathbb{E}_{p(x_1, x_2, y)}[-\log q_\theta(y\mid x_1, x_2)] \nonumber \\
&= C_r
- \mathbb{E}_{p(\tilde{x}_1, x_2)}\, D_{\mathrm{KL}}\!\left(q_\theta(\cdot\mid \tilde{x}_1, x_2)\,\|\,r(\cdot)\right)
- \mathbb{E}_{p(x_1, x_2, y)}[-\log q_\theta(y\mid x_1, x_2)].
\label{eq:mipd-surrogate}
\end{align}
Up to the additive constant $C_r$, maximizing $\widehat{\mathrm{MIPD}}_1$ is equivalent to minimizing the SynIB loss: jointly minimizing the cross-entropy on intact inputs and the KL-to-reference on corrupted inputs.

\paragraph{Properties of the surrogate.}
Three observations clarify how $\widehat{\mathrm{MIPD}}_1$ relates to the true $\mathrm{MIPD}_1$.

First, $\widehat{\mathrm{MIPD}}_1 = \mathrm{MIPD}_1$ whenever $q_\theta(\cdot\mid x_1, x_2) = p(\cdot\mid x_1, x_2)$ and $q_\theta(\cdot\mid \tilde{x}_1, x_2) = p(\cdot\mid \tilde{x}_1, x_2)$. In this case both the upper bound in Eq.~\eqref{eq:ba-upper} and the surrogate substitution in Eq.~\eqref{eq:H-surrogate} are tight.

Second, the cross-entropy term in $\widehat{\mathrm{MIPD}}_1$ is itself the optimization signal that drives $q_\theta(\cdot\mid x_1, x_2)$ toward $p(\cdot\mid x_1, x_2)$, so the surrogate is self-correcting on the intact-input branch: improvements in $q_\theta$ tighten the bound in Eq.~\eqref{eq:ba-upper}.

Third, the surrogate substitution in Eq.~\eqref{eq:H-surrogate} is not a one-sided bound: the gap $\widehat{H}_\theta(Y\mid \tilde{X}_1, X_2) - H(Y\mid \tilde{X}_1, X_2) = \mathbb{E}_{p(\tilde{x}_1, x_2)}[D_{\mathrm{KL}}(p\,\|\,r) - D_{\mathrm{KL}}(q_\theta\,\|\,r)]$ can take either sign depending on the relative position of $q_\theta$ and $p$ with respect to $r$. The surrogate should therefore be understood as a tractable target whose sign-correct alignment with $\mathrm{MIPD}_1$ relies on $q_\theta$ approximating $p$ in distribution, which the cross-entropy term promotes during training.

\subsection{MIPD as a Proxy for CMI}
\label{app:mipd-cmi-proxy}
Beyond serving as the optimization target, $\mathrm{MIPD}_1$ admits a clean information-theoretic interpretation as a lower-bounded proxy function for the conditional mutual information $I(X_1; Y\mid X_2)$. Expanding the definition of MIPD in terms of conditional mutual informations gives $\mathrm{MIPD}_1 = I(X_1; Y\mid X_2) - I(\tilde{X}_1; Y\mid X_2)$, and since $I(\tilde{X}_1; Y\mid X_2) \ge 0$ for any random variable $\tilde{X}_1$, it follows directly that $\mathrm{MIPD}_1 \le I(X_1; Y\mid X_2)$, with the gap given exactly by the residual conditional information retained after perturbation:
\begin{equation}
I(X_1; Y\mid X_2) - \mathrm{MIPD}_1 = I(\tilde{X}_1; Y\mid X_2).
\label{eq:gap-equals-residual}
\end{equation}
This identity has a clear interpretation: MIPD recovers the full CMI exactly when the perturbation $\tilde{X}_1$ is conditionally uninformative about $Y$ given $X_2$, i.e., when $Y \perp\!\!\!\perp \tilde{X}_1 \mid X_2$. In this case, all task-relevant information that $X_1$ contributed beyond $X_2$ has been removed by the corruption, and the drop in predictability $\mathrm{MIPD}_1$ equals the conditional information $X_1$ originally carried. More generally, if the corruption leaves residual information $I(\tilde{X}_1; Y\mid X_2) \le \varepsilon$, then MIPD approximates CMI to within $\varepsilon$.

The proxy quality thus depends entirely on how thoroughly the perturbation removes task-relevant information from $X_1$ beyond what $X_2$ already provides. Designing $\tilde{X}_1$ to make this residual small motivates the masking constructions in Sec.~\ref{sec:masking}: random masking corrupts coordinates uniformly, while learned masking targets the features the unimodal predictor relies on, both aiming to drive $I(\tilde{X}_1; Y\mid X_2)$ as close to zero as possible while preserving the rest of the distribution.
\section{Variational Families and Closed-Form KL Divergences}
\label{app:variational}

The SynIB objective requires evaluating $-\log q_\theta(y \mid x_1, x_2)$ for the cross-entropy term and $D_{\mathrm{KL}}(q_\theta(\cdot \mid \tilde{x}_1, x_2) \,\|\, r(\cdot))$ for the KL term. Both quantities admit closed-form expressions for standard variational families, removing the need for stochastic estimators. We use the same family for both intact and corrupted inputs, with the conditioning input being the only difference between the two passes.

\subsection{Gaussian Variational Family}
For continuous targets $Y \in \mathbb{R}^d$, we model the variational predictive distribution as $q_\theta(y \mid \tilde{x}_1, x_2) = \mathcal{N}(y \mid \mu_\theta, \Sigma_\theta)$, with reference distribution $r(y) = \mathcal{N}(0, I)$. The mean $\mu_\theta$ and covariance $\Sigma_\theta$ are outputs of the variational decoder and depend on $(\tilde{x}_1, x_2)$. The KL divergence admits the closed form
\begin{equation}
D_{\mathrm{KL}}(q_\theta \,\|\, r)
= \tfrac{1}{2}\!\left(\|\mu_\theta\|_2^2 + \mathrm{tr}\,\Sigma_\theta - \log\det\Sigma_\theta - d\right),
\end{equation}
which is minimized when $q_\theta = r$ (i.e., $\mu_\theta = 0$, $\Sigma_\theta = I$) and grows with both the magnitude of $\mu_\theta$ and the deviation of $\Sigma_\theta$ from the identity, providing a differentiable uncertainty penalty under modality corruption.

\subsection{Categorical Variational Family}
For categorical targets $Y \in \{1, \dots, K\}$, we model the variational predictive distribution as $q_\theta(y \mid \tilde{x}_1, x_2) = \mathrm{Categorical}(y; \pi_\theta)$, with reference $r(y) = \mathrm{Categorical}(y; \rho)$, where $\pi_\theta = \pi_\theta(\tilde{x}_1, x_2) \in \Delta^{K-1}$ is the output of the variational decoder and $\rho \in \Delta^{K-1}$ is a fixed reference distribution (we use $\rho_k = 1/K$, the uniform prior, in all experiments). The KL divergence admits the closed form
\begin{equation}
D_{\mathrm{KL}}(q_\theta \,\|\, r)
= \sum_{k=1}^{K} \pi_{\theta,k} \log \frac{\pi_{\theta,k}}{\rho_k}.
\end{equation}
The penalty is minimized when $\pi_\theta = \rho$ and grows as $\pi_\theta$ concentrates on any single class, discouraging confident predictions under modality corruption. The binary case $K = 2$ recovers the Bernoulli form $D_{\mathrm{KL}}(q_\theta \,\|\, r) = \pi_\theta \log(\pi_\theta / \rho) + (1 - \pi_\theta) \log((1 - \pi_\theta) / (1 - \rho))$, used for tasks with binary targets.
\section{Masking Operator Instantiations}
\label{app:masking}

This appendix specifies the concrete realizations of the masking operator $\mathcal{T}(\cdot;\,M)$ introduced in Sec.~\ref{sec:masking}. All variants implement the same semantics, selectively corrupting a subset of representation coordinates while leaving the rest intact, and differ only in how the mask $M$ is constructed and how the suppression is realized within a given architecture. Throughout this appendix we follow the convention from the main text: $M_j = 1$ denotes a corrupted coordinate, $M_j = 0$ a preserved one.

\paragraph{Feature replacement via gating.}
In non-attention-based fusion networks, masking is implemented by replacing corrupted coordinates with noise through the feature-wise gate
\begin{equation}
\tilde{x}_1 = (1 - M) \odot x_1 + M \odot \epsilon,
\label{eq:gate-app}
\end{equation}
where $M \in \{0,1\}^{d_1}$ and $\epsilon$ is sampled independently of the target $Y$. The construction preserves dimensionality and, by design, removes the task-relevant signal carried by the corrupted features while leaving the rest of the input distribution intact. The same expression applies when masking is performed in the latent space: substituting $z_1 = f_{\theta_{c_1}}(x_1)$ for $x_1$ yields the corresponding latent-space gate.

\paragraph{Oracle masking.}
In synthetic settings where the synergistic coordinates of $X_1$ are known by construction, we define an oracle mask
\begin{equation}
M^\star_j = \mathbf{1}[j \in S_{\mathrm{syn}}],
\end{equation}
which corrupts exactly the designated synergistic features and preserves everything else. Applied through Eq.~\eqref{eq:gate-app}, this isolates the contribution of synergy by removing only the coordinates whose information is recoverable from the joint $(X_1, X_2)$ but not from either modality alone. Oracle masking is used exclusively for diagnostic experiments and serves as an upper-bound reference point: it characterizes how SynIB would behave if it could perfectly identify synergistic features, without claiming that this information is available in real-world settings.

\paragraph{Random masking.}
For stochastic masking, each coordinate is independently corrupted according to $M_j \sim \mathrm{Bernoulli}(\pi)$, where $\pi \in (0,1)$ controls the expected corruption level. Random masking makes no assumption about which coordinates carry which type of information, providing a task-agnostic baseline that removes signal uniformly across dimensions and surfaces synergy whenever any corrupted subset breaks the joint prediction.

To avoid introducing out-of-distribution artifacts when corrupting, the noise $\epsilon$ in Eq.~\eqref{eq:gate-app} is sampled to match the marginal statistics of the representation being corrupted. Let $z_1 = f_{\theta_{c_1}}(x_1)$ and maintain exponential moving averages over minibatches $\mathcal{B}$:
\begin{equation}
\begin{aligned}
\mu_t
&= \alpha \mu_{t-1} + (1-\alpha)\,\mathbb{E}_{\mathcal{B}}[z_1], \\
\sigma_t^2
&= \alpha \sigma_{t-1}^2 + (1-\alpha)\,\mathbb{E}_{\mathcal{B}}\!\left[(z_1 - \mathbb{E}_{\mathcal{B}}[z_1])^2\right].
\end{aligned}
\end{equation}
Noise is then sampled as $\epsilon \sim \mathcal{N}(\mu_t, \mathrm{diag}(\sigma_t^2))$. The momentum $\alpha$ controls the smoothness of the running statistics; we use $\alpha = 0.99$ in all experiments. To prevent the model from exploiting structure in the replacement distribution, the same noise process is applied to fully corrupted counterfactual inputs during training.

\paragraph{Learned masking.}
The learned mask $M_{\psi_1}$ introduced in Sec.~\ref{sec:masking} flags the minimum set of coordinates whose corruption collapses unimodal prediction. Training $M_{\psi_1}$ adversarially against the unimodal predictor $f_{\theta_{c_1}}$ via Eq.~\eqref{eq:mask-loss} therefore identifies the unimodally important features of $X_1$. The SynIB counterfactual then uses the complement $\bar{M}_{\psi_1} = \mathbf{1} - M_{\psi_1}$ as its corruption mask in Eq.~\eqref{eq:gate-app}, leaving the unimodally important coordinates intact and corrupting the rest. Confidence under this corruption signals reliance on cues that survive the removal of $X_1$'s non-unimodal signal, which the SynIB KL term penalizes. The mask network $M_{\psi_1}$ is parameterized as a small MLP producing logits $\ell \in \mathbb{R}^{d_1}$, with binary samples drawn through a Gumbel-softmax relaxation~\citep{jang2016categorical} to enable gradient flow during joint training.

\section{Likelihood Perspective on Learned Masking}
\label{app:density}

Learned masking can be interpreted as a structured perturbation of the conditional likelihood implicitly defined by the unimodal predictor $f_{\theta_{c_1}}$. This appendix makes the connection explicit and clarifies how the learned mask differs from the fixed alternatives (random, oracle).

Cross-entropy training of $f_{\theta_{c_1}}$ implicitly defines a conditional likelihood $p_{\theta_{c_1}}(y \mid z_1)$ over the target given the modality-1 representation, with the cross-entropy loss equal to its negative log-likelihood under the data distribution:
\begin{equation}
\mathrm{CE}\!\left(f_{\theta_{c_1}}(z_1),\, y\right)
= -\mathbb{E}_{p(y, z_1)}\!\left[\log p_{\theta_{c_1}}(y \mid z_1)\right].
\end{equation}
Perturbations of the representation that increase cross-entropy therefore correspond directly to degradations of this conditional log-likelihood.

The learned mask $M_{\psi_1}$ exploits this relationship by seeking the sparsest perturbation that maximally degrades the likelihood. With $\tilde{Z}_1 = \mathcal{T}(Z_1;\, M_{\psi_1})$ denoting the gated representation under the new convention ($M_{\psi_1, j} = 1$ corrupts coordinate $j$, $M_{\psi_1, j} = 0$ preserves it), the mask training objective in Eq.~\eqref{eq:mask-loss} can be equivalently written as
\begin{equation}
\psi_1^\star
= \arg\max_{\psi_1}\;
\mathbb{E}_{p(y, z_1)}\!\left[\log \frac{p_{\theta_{c_1}}(y \mid Z_1)}{p_{\theta_{c_1}}(y \mid \tilde{Z}_1)}\right]
- \lambda_M \|M_{\psi_1}\|_1,
\end{equation}
where the first term measures the average log-likelihood drop induced by the perturbation and the second imposes an $\ell_1$ relaxation of an $\ell_0$ sparsity constraint on the corruption set. Solutions concentrate corruption on the minimal subset of representation coordinates whose removal most degrades the unimodal likelihood, identifying the features $f_{\theta_{c_1}}$ relies on for prediction.

This view places the three masking strategies on a common axis. Random masking ($M_j \sim \mathrm{Bernoulli}(\pi)$) corrupts coordinates uniformly without consulting the likelihood. Oracle masking corrupts a fixed set $S_{\mathrm{syn}}$ defined by ground-truth synergy structure available only in synthetic settings. Learned masking adaptively concentrates corruption on the coordinates that most reduce the likelihood of $f_{\theta_{c_1}}$, which we then complement in the SynIB counterfactual to leave only the unimodally-important features intact.
\section{Additional Synthetic Experiment Details}
\label{app:synthetic_details}

We use two synthetic XOR datasets, each targeting a different aspect of SynIB's behavior. Sec.~\ref{app:spurious_xor} (\emph{Spurious XOR}) tests robustness when one modality contains a label-correlated shortcut whose strength sweeps continuously between absent and dominant. Sec.~\ref{app:pid_xor} (\emph{PID-Controlled XOR}) generates examples whose label is determined by a single PID source per sample, allowing the dataset's mixture of unique, redundant, and synergistic information to be controlled directly. Both datasets use the modality-indexing convention from the main text ($X_1, X_2$) and the masking convention $M_j = 1$ corrupts, $M_j = 0$ preserves.

\subsection{Spurious XOR}
\label{app:spurious_xor}

\subsubsection{Task and Data Generation}

The Spurious XOR task is a bimodal binary classification problem in which neither modality alone is informative about $y$, but their XOR is. Latent bits $b_1, b_2 \sim \mathrm{Bernoulli}(0.5)$ are sampled independently and the label is set to $y = b_1 \oplus b_2$, with modality $X_m$ encoding bit $b_m$. Modality $X_1$ additionally contains a spurious feature whose correlation with $y$ during training is controlled by a parameter $\beta$ and is removed at test time, so a model that latches onto the shortcut fails to generalize.

Both modalities are 64-dimensional and partitioned into disjoint coordinate blocks. Modality 1 contains a 4-coordinate signal block at $[0, 4)$ encoding $b_1$, a 6-coordinate spurious block at $[4, 10)$ correlated with $y$ at training time, and 54 pure-noise coordinates at $[10, 64)$. Modality 2 contains a 4-coordinate signal block at $[0, 4)$ encoding $b_2$ and 60 pure-noise coordinates at $[4, 64)$. Pure-noise coordinates are sampled as $\mathcal{N}(0, 1)$.

Both signal blocks are mean-shifted Gaussians, $X_m^{\mathrm{sig}} = \mu \cdot (2 b_m - 1) \cdot \mathbf{1}_4 + \sigma_{\mathrm{sig}}\,\varepsilon$ with $\varepsilon \sim \mathcal{N}(0, I_4)$, $\mu = 2.2$, and $\sigma_{\mathrm{sig}} = 0.55$. The spurious block in $X_1$ is a continuous mean-shifted Gaussian whose label-correlation strength is controlled by $\beta \geq 0$:
\begin{equation}
X_1^{\mathrm{spur}} = 1.3 \cdot \left[\,\beta \cdot (2 y_{\mathrm{eff}} - 1) \cdot \mathbf{1}_6 + \sqrt{\max(0, 1 - \beta^2)} \cdot \varepsilon\,\right], \quad \varepsilon \sim \mathcal{N}(0, I_6),
\end{equation}
with $y_{\mathrm{eff}} = y$ at training time and $y_{\mathrm{eff}} = 0$ at test time. The parameter $\beta$ is a continuous SNR knob, not a probability: at $\beta = 0$ the spur is pure noise; at $\beta = 1$ the noise term vanishes and the spur becomes deterministic in $y$; the $\max(0, \cdot)$ clamp permits values $\beta > 1$, which further amplify the deterministic mean-shift while keeping the noise term zero. We sweep $\beta \in \{0, 0.2, 0.4, 0.6, 0.8, 1.0, 2.0\}$. The training set contains 300 samples, the validation set 120, and the test set 30{,}000; the small training set is deliberate and gives the spurious shortcut a real chance to dominate the cross-modal XOR signal during fitting.

\subsubsection{Model, Objective, and Mask Variants}

The fusion network has modality-specific encoders, two unimodal heads, and a fusion head. Each encoder $f_{\theta_m}$ is a single hidden layer, $\mathrm{Linear}(64 \to 16) \to \mathrm{ReLU} \to \mathrm{Dropout}$. Unimodal heads are $\mathrm{Linear}(16 \to 1)$, trained against $y$ with weight $\lambda_{\mathrm{uni}} = 0.05$. The fusion head concatenates the two encoder outputs and applies $\mathrm{Linear}(32 \to 32) \to \mathrm{ReLU} \to \mathrm{Dropout} \to \mathrm{Linear}(32 \to 1)$, producing a single logit trained with binary cross-entropy. SynIB regularizes the fusion logit symmetrically across modalities,
\begin{equation}
\mathcal{L} = \mathcal{L}_{\mathrm{CE}} + \lambda \sum_{i \in \{1, 2\}} D_{\mathrm{KL}}\!\left(P_f(\cdot \mid \tilde{X}_i, X_{\neg i}) \,\big\|\, \mathrm{Bernoulli}(0.5)\right),
\end{equation}
where $X_{\neg i}$ denotes the unmasked modality. The reference distribution is uniform over the binary outcomes; only the fusion logit is regularized, not the unimodal heads.

Corrupted coordinates are replaced via $\tilde{x} = (1 - M) \odot x + M \odot \varepsilon$ with $\varepsilon \sim \mathcal{N}(0, I)$. The oracle mask $M^\star$ flags the signal blocks $[0, 4)$ of both modalities (the coordinates encoding $b_1, b_2$); the spurious block in $X_1$ is not flagged, so any confident prediction under oracle corruption would have to rely on the shortcut. Random masking corrupts each coordinate of each modality independently with probability $\pi = 0.4$. Learned masking uses per-example, per-coordinate logits initialized at $-4$, with an inner loop of 50 Adam steps (lr $=3.0$, sparsity $\lambda_M = 0.03$, temperature $\tau = 1.0$) finding the smallest mask whose corruption breaks the unimodal head; the discovered mask is hard-thresholded at $0.5$, and the SynIB pass corrupts the complement, leaving the unimodally-important coordinates intact.

\subsubsection{Training Configuration}

\begin{figure}[t]
    \centering
    \includegraphics[width=\linewidth]{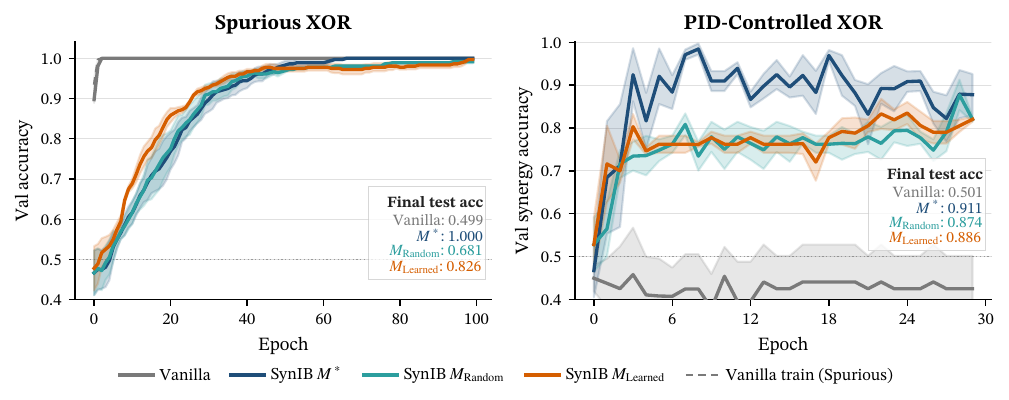}
  \caption{
\textbf{Validation accuracy over training on both synthetic benchmarks.} Mean $\pm$ standard error across three seeds. \textbf{Left}: Spurious XOR ($\beta = 1.0$). All four methods reach $\approx 1.0$ on the in-distribution validation set, fitting the training distribution equally well. The gray dashed curve shows training accuracy for vanilla fusion, which saturates within the first epoch as the model locks onto the shortcut. The annotated final out-of-distribution test accuracies reveal the divergence: vanilla collapses to chance ($0.50$) while oracle masking recovers fully ($1.00$); learned masking reaches $0.83$ and random masking $0.68$. Random masking shows the largest seed-to-seed variance (per-seed test accuracies $0.58$--$0.78$), reflecting sensitivity to which coordinates the random mask happens to corrupt. \textbf{Right:} PID-Controlled XOR. Synergy-restricted validation accuracy at simplex point $(0.45, 0, 0.45, 0.10)$. Vanilla plateaus at chance on the synergy slice; oracle, learned, and random masking each rise above $0.85$, with learned masking ($0.89$) tracking oracle ($0.91$) within three percentage points.}  
  \label{fig:synergy-acc-over-training}
\end{figure}

We train with Adam, batch size 128, for 100 epochs, across seeds $\{0, 1, 2\}$ and dropout $0$. The baseline uses learning rate $10^{-2}$ and weight decay $10^{-3}$; the SynIB variants use learning rate $3 \times 10^{-4}$ and weight decay $10^{-5}$, with regularization strength $\lambda = 10$. Validation accuracy curves over training are reported in Figure~\ref{fig:synergy-acc-over-training} (left panel) with final test accuracies. Reported numbers are mean $\pm$ standard error across the 3 seeds.

\subsection{PID-Controlled XOR}
\label{app:pid_xor}

\subsubsection{Task and Data Generation}

The PID-Controlled XOR task is a bimodal binary classification problem in which each sample's label is generated by a single PID source: unique to modality 1 ($U_1$), unique to modality 2 ($U_2$), redundant ($R$), or synergistic ($S$). The dataset's mixture is set by a categorical distribution $p(A) = (p_{U_1}, p_{U_2}, p_R, p_S)$, allowing direct control over the prevalence of each information type. For Figure~\ref{fig:pid_xor} we hold $p_{U_2} = 0$ and sweep the remaining three coordinates over the probability simplex on a regular grid with step $0.05$.

Both modalities are 32-dimensional and partitioned into four disjoint blocks of 6, 6, 6, and 14 coordinates. The first block carries the modality's unique signal ($U_1$ for modality 1, $U_2$ for modality 2), the second block carries the redundant signal, the third block carries the synergistic signal, and the fourth block is pure noise. Block coordinate sets are fixed across samples.

Each sample is generated by initializing all coordinates of $X_1, X_2$ as i.i.d.\ $\mathcal{N}(0, 1)$, sampling $y \sim \mathrm{Bernoulli}(0.5)$ independently, sampling an active source $A \sim p(A)$, overwriting the block coordinates of the active source with label-aligned signal as described below, and finally standardizing each coordinate using the training-set mean and standard deviation. Blocks not corresponding to the active source remain at their initial $\mathcal{N}(0, 1)$ noise.

Three pairs of i.i.d.\ random projection matrices $P_{u_m}, P_{r_m}, P_{s_m} \in \mathbb{R}^{d_m \times 4}$ are sampled once at script load (seeded for reproducibility) with entries i.i.d.\ $\mathcal{N}(0, 0.5^2)$. Sign-aligned latents are constructed by sampling $z \sim \mathcal{N}(0, s^2 I_4)$ for the appropriate source-specific scale $s$ and applying $z \leftarrow (2 b - 1) \cdot |z|$ elementwise for the relevant bit $b$. The full projected vector is computed and only the corresponding block coordinates are written back into the modality. We use $s_u = s_r = s_s = 3$.

When source $U_m$ is active, $z$ is sampled with $b = y$ and $X_m[\mathrm{unique}] = (P_{u_m} z)[\mathrm{unique}]$; the other modality's unique block remains noise. When source $R$ is active, a single shared latent $z_R$ is sign-aligned with $y$ and projected through different matrices $P_{r_1}, P_{r_2}$ into the redundant blocks of both modalities, so the two redundant blocks are correlated through $z_R$ but not identical. When source $S$ is active, a latent bit $b \sim \mathrm{Bernoulli}(0.5)$ is sampled with $b_1 = b$ and $b_2 = b \oplus y$ so that $b_1 \oplus b_2 = y$; independent latents $z^{(m)}$ are sign-aligned with $b_m$ and projected into modality $m$'s synergistic block, making each modality alone uninformative about $y$ while their XOR recovers it. The training set contains 1800 samples per simplex point, with 200 held out for validation and 4000 for test.

\subsubsection{Model, Objective, and Mask Variants}

The PID-XOR model uses substantially more capacity than the Spurious-XOR model, reflecting the harder task of distinguishing four signal types across more coordinates. Each encoder is two hidden layers, $\mathrm{Linear}(32 \to 1024) \to \mathrm{ReLU} \to \mathrm{Dropout} \to \mathrm{Linear}(1024 \to 1024) \to \mathrm{ReLU}$. Unimodal heads are $\mathrm{Linear}(1024 \to 1024) \to \mathrm{ReLU} \to \mathrm{Linear}(1024 \to 1)$, with the unimodal-head loss entering the total objective with weight $\lambda_{\mathrm{uni}} = 1.0$. The fusion head concatenates the two encoder outputs (2048-d) and applies $\mathrm{Linear}(2048 \to 128) \to \mathrm{ReLU} \to \mathrm{Dropout} \to \mathrm{Linear}(128 \to 128) \to \mathrm{ReLU} \to \mathrm{Linear}(128 \to 1)$. The SynIB objective takes the same form as in Sec.~\ref{app:spurious_xor}, with $\mathrm{Bernoulli}(0.5)$ as the reference and KL applied to the fusion logit.

The same gating mechanism is used, with replacement noise $\mathcal{N}(0, 1)$ per coordinate. The oracle mask $M^\star$ corrupts only the synergistic blocks of both modalities; unique, redundant, and noise blocks are preserved. Random masking corrupts each coordinate of each modality independently with probability $\pi = 0.5$. Learned masking uses an inner loop of 20 Adam steps (lr $=0.1$, $\tau = 1.0$); the hard binary mask discovered by the inner loop is used as the corruption pattern in the outer SynIB pass.

\section{Disentangling Masking Effects from SynIB}
\label{app:masking_vs_synib}

An arising question is whether the gains from SynIB stem from the proposed information-theoretic objective or simply from exposing the model to masked inputs during training. To rule out the latter, we compare SynIB against a masking-only baseline that uses the same mask-construction mechanism as SynIB but applies it as input augmentation rather than as a counterfactual KL penalty: each batch is augmented with masked copies of the original samples, and training proceeds with cross-entropy on the union of intact and masked inputs.

We instantiate two variants of this baseline that mirror SynIB's mask choices: \emph{masked input with oracle} corrupts the unique and redundant blocks (preserving the synergistic block) so the model is exposed to inputs that carry only synergistic signal, and \emph{masked input with random masking} applies a random Bernoulli mask matching the SynIB-random configuration.

\begin{figure*}[t]
\centering
\includegraphics[width=\textwidth]{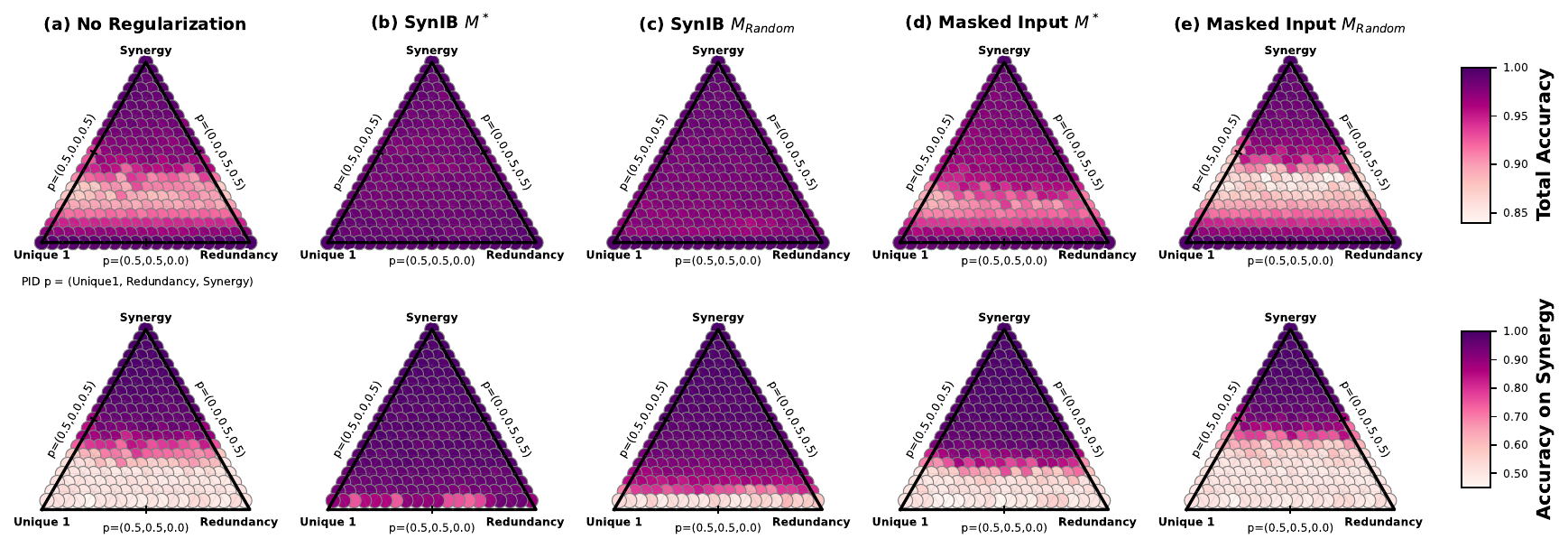}
\caption{Masking-based baselines vs SynIB across PID compositions. Columns: (a) no regularization, (b) SynIB with oracle masking $M^\star$, (c) SynIB with random masking, (d) masked-input training with oracle, (e) masked-input training with random masking. Triangles sweep the PID simplex over $(p_{U_1}, p_R, p_S)$. Top row: total accuracy; bottom row: accuracy restricted to synergistic samples. Masked-input baselines yield limited gains, while SynIB recovers synergistic structure across the simplex.}
\label{fig:masking_vs_synib}
\end{figure*}

Figure~\ref{fig:masking_vs_synib} reports the comparison across the PID simplex. The masking-only baselines yield limited improvements over no regularization and fail to recover the synergistic structure, particularly when synergy is dominant. SynIB, in contrast, achieves substantially higher accuracy across the simplex and on the synergy-restricted subset. The gap indicates that input augmentation alone is insufficient: SynIB's gains depend on explicitly penalizing confident predictions under modality corruption, not on data exposure to masked inputs.

\section{Synergy-Driven Irony Dataset Construction}
\label{app:irony_dataset}

This appendix provides the construction details for the synthetic irony benchmark introduced in Sec.~\ref{sec:multibench}, along with the corresponding model architecture, training setup, and evaluation protocol.

\subsection{Base Dataset and Splits}

We start from CREMA-D~\citep{cao2014crema}, a crowd-sourced audio--visual emotion recognition dataset of short clips in which actors speak sentences with a target emotion. Following the standard 6-class taxonomy, we use neutral (NEU), happy (HAP), sad (SAD), fearful (FEA), disgusted (DIS), and angry (ANG); we add a synthetic 7th class for irony. We use a 5-fold speaker-disjoint split with no actor appearing in more than one of (train, val, test) within any fold. After dropping samples missing audio, video, or face crops, each fold contains 5{,}820 samples (approximately 4{,}043 train / 878 val / 899 test); base classes are mildly imbalanced with HAP largest and SAD smallest at a roughly $1.5\times$ ratio.

\subsection{Mutation Procedure}

Let $D_{\text{original}} = \{(V_i, A_i, L_i)\}_{i=1}^N$ denote the original set of (video, audio, label) triplets, and let $\bar{N}$ denote the mean number of samples per base class within a given split. The irony rate $\alpha$ controls the number of mutated samples, $|I| = \lfloor \alpha \bar{N} \rfloor$, drawn uniformly without replacement from the base-class pool of that split. For each selected index $i$ with base label $L_i$, we sample a donor index $j$ from a designated set of contradiction labels and form a mutated triplet $(V_i, A_j, L_{\text{irony}})$. The original audio is replaced by the donor's audio while video and identity are preserved; the new label is the irony class. The resulting dataset is
\begin{equation}
D_{\text{new}} = \{(V_i, A_j, L_{\text{irony}})\}_{i \in I} \cup \{(V_i, A_i, L_i)\}_{i \notin I}.
\end{equation}
Mutation is in-place: each mutated sample replaces its original, so the total dataset size is preserved and the irony class population is controlled solely by $\alpha$. We sweep $\alpha \in \{0.1, 0.3, 0.5, 0.8, 1.0, 2.0\}$.

The donor set for each base class is fixed by a semantic contradiction map: HAP is paired with one of $\{$SAD, FEA, DIS, ANG$\}$, while SAD, FEA, DIS, and ANG are each paired with HAP. NEU samples have no semantically opposite class in this taxonomy and are paired with any non-matching label uniformly at random. This map ensures that ironic pairings produce strong cross-modal contradictions rather than subtle disagreements between similar emotions. Donors are drawn from the same split as the mutated sample (train donors for train mutations, etc.) to prevent identity leakage across train and test, and the same $\alpha$ is applied to all three splits. Each mutated sample is, by construction, individually consistent with both its visual base emotion and its donor audio's emotion: neither modality alone identifies the sample as ironic, so the irony label is recoverable only from the contradiction between modalities.

\subsection{Feature Representations}

Audio is represented as a log-magnitude short-time Fourier transform (STFT) spectrogram. Waveforms are resampled to 22{,}050 Hz, tiled and cropped to 3 seconds, and clipped to $[-1, 1]$ before computing an STFT with $n_{\text{FFT}} = 512$ and hop length 353. The resulting spectrogram is $\log(|\cdot| + 10^{-7})$, producing a single-channel $257 \times 188$ representation.

Video is represented by 3 RGB frames sampled at 1 fps, resized to $224 \times 224$, and normalized using ImageNet statistics. Training-time augmentation uses random resized crops and random horizontal flips; validation and test use deterministic resizing.

\subsection{Model Architecture}

Each modality is encoded by a ResNet18 backbone adapted for its input shape. The audio backbone takes the single-channel spectrogram and applies a global average pool to produce a 512-dimensional representation; the video backbone takes the 3-frame clip and applies a 3D global average pool to produce a 512-dimensional representation. Both backbones are pretrained unimodally on CREMA-D before joint training, with the same pretrained checkpoints used as initialization across all methods.

The fusion trunk concatenates the two 512-dimensional encoder outputs and applies $\mathrm{Linear}(1024 \to 64) \to \mathrm{ReLU} \to \mathrm{Dropout}(0.1)$. The fusion head is a single $\mathrm{Linear}(64 \to 7)$ producing the 7-way classification logits. Auxiliary unimodal heads ($\mathrm{Linear}(512 \to 64) \to \mathrm{ReLU} \to \mathrm{Dropout}(0.1) \to \mathrm{Linear}(64 \to 7)$) are attached to each encoder for the SynIB inner loop. All baselines share the same ResNet18 backbones and pretrained initialization; they differ in the fusion mechanism according to each method's specification.

\subsection{Training and SynIB Configuration}

We train with Adam ($\beta_1 = 0.9$, $\beta_2 = 0.999$), learning rate $10^{-5}$, weight decay $10^{-5}$, and batch size 64, with cosine annealing and a 780-step warmup. Maximum training length is 1500 epochs with early stopping on validation accuracy (patience 30 epochs). The classification loss is unweighted cross-entropy across all 7 classes, with no per-class reweighting.

For SynIB, we use random masking on each modality with corruption probability $\pi$ and reference distribution given by the complementary modality's unimodal prediction, so the KL term penalizes the fused prediction for departing from what the unmasked modality alone would predict. The regularization strength is $\lambda$. SynIB-specific values are held constant across all $\alpha$ for a given run.

\subsection{Evaluation}

Models are evaluated on the test split of each fold using two metrics: macro F1 across all 7 classes (reported as Total F1 in Figure~\ref{fig:crema_irony_rates}) and per-class F1 for the irony class (reported as Irony F1). Both metrics are computed on the full test set, which contains $\lfloor \alpha \bar{N}_{\text{test}} \rfloor$ irony samples generated by the same mutation procedure used at training time. Reported values in Figure~\ref{fig:crema_irony_rates} are mean and standard deviation across 3 folds.

\begin{figure}[t]
    \centering
    \includegraphics[width=\linewidth]{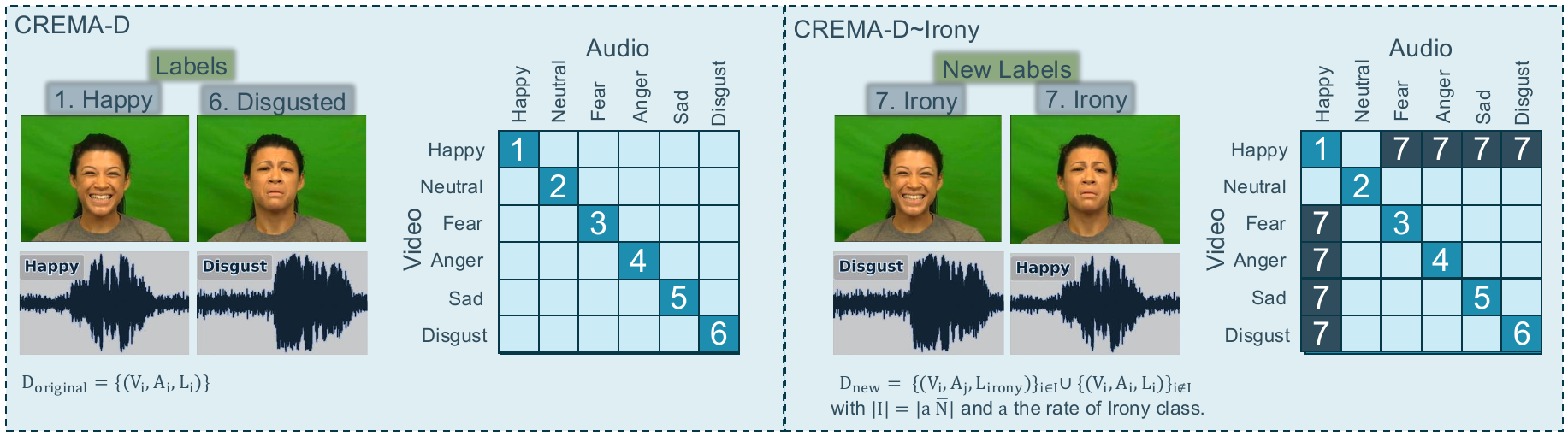}
    \caption{Construction of the synthetic irony class. Each ironic sample is built by pairing a video $V_i$ from one base emotion (e.g., happy) with a donor audio $A_j$ from a contradicting emotion (e.g., disgust). The new label is irony. The dataset is $D_{\text{new}} = \{(V_i, A_j, L_{\text{irony}})\}_{i \in I} \cup \{(V_i, A_i, L_i)\}_{i \notin I}$, with $|I| = \lfloor \alpha \bar{N} \rfloor$ controlled by the irony-rate hyperparameter $\alpha$. Neither modality alone identifies an ironic sample; the label requires reading the cross-modal contradiction.}
    \label{fig:data_generation}
\end{figure}

\subsection{Numerical Results}
\label{app:crema_irony_full}

Table~\ref{table:perf_results_irony_full} reports the per-method F1 scores summarized in Figure~\ref{fig:crema_irony_rates}, with standard deviations across folds. Each cell shows irony-class F1 over total F1 (mean $\pm$ standard deviation across 3 folds); bold marks the best entry per column for each metric. Uni-Video and Uni-Audio do not predict the irony class and report total F1 only. Reported values cover the four irony rates used in the main figure ($\alpha \in \{0.1, 0.5, 1.0, 2.0\}$).

\begin{table}[ht]
\centering
\caption{F1 scores on the CREMA-D irony recognition task under varying irony rates $\alpha$, reported as irony-class F1 / total F1 (mean $\pm$ standard deviation across 3 folds). Bold marks the best per column for each metric. Dashes indicate that the corresponding metric does not apply.}
\label{table:perf_results_irony_full}
\small
\setlength{\tabcolsep}{4pt}
\renewcommand{\arraystretch}{1.15}
\begin{tabular}{lcccc}
\toprule
Irony rate $\alpha$ & 0.1 & 0.5 & 1.0 & 2.0 \\
\midrule
Uni-Video       & ---\,/\,41.0{\tiny$\pm$3.5}     & ---\,/\,39.9{\tiny$\pm$2.9}     & ---\,/\,38.3{\tiny$\pm$3.7}     & ---\,/\,28.0{\tiny$\pm$14.7}    \\
Uni-Audio       & ---\,/\,49.8{\tiny$\pm$1.7}     & ---\,/\,46.9{\tiny$\pm$2.4}     & ---\,/\,45.7{\tiny$\pm$1.3}     & ---\,/\,39.6{\tiny$\pm$1.2}     \\
\midrule
Ensemble        & 3.0{\tiny$\pm$5.2}\,/\,59.7{\tiny$\pm$4.3}    & 15.0{\tiny$\pm$8.0}\,/\,58.4{\tiny$\pm$2.8}    & 36.8{\tiny$\pm$6.3}\,/\,59.6{\tiny$\pm$3.2}  & 54.3{\tiny$\pm$3.8}\,/\,38.3{\tiny$\pm$21.4} \\
Vanilla fusion  & 6.4{\tiny$\pm$5.6}\,/\,59.3{\tiny$\pm$3.2}    & 18.3{\tiny$\pm$12.2}\,/\,57.4{\tiny$\pm$3.9}   & 39.7{\tiny$\pm$0.2}\,/\,58.2{\tiny$\pm$1.0}  & 50.0{\tiny$\pm$5.5}\,/\,54.5{\tiny$\pm$4.6}           \\
\midrule
D\&R            & 0.0{\tiny$\pm$0.0}\,/\,55.1{\tiny$\pm$1.4}    & 9.0{\tiny$\pm$6.1}\,/\,55.7{\tiny$\pm$1.6}     & 25.8{\tiny$\pm$9.0}\,/\,54.3{\tiny$\pm$1.6}           & 45.2{\tiny$\pm$5.9}\,/\,47.0{\tiny$\pm$0.7}           \\
MMPareto        & 3.7{\tiny$\pm$6.4}\,/\,58.9{\tiny$\pm$2.3}    & 13.0{\tiny$\pm$4.2}\,/\,58.6{\tiny$\pm$0.6}    & 20.0{\tiny$\pm$0.8}\,/\,55.5{\tiny$\pm$1.1}           & 41.5{\tiny$\pm$2.6}\,/\,50.2{\tiny$\pm$1.8}           \\
ReconBoost      & 0.0{\tiny$\pm$0.0}\,/\,57.6{\tiny$\pm$3.1}    & 11.6{\tiny$\pm$4.4}\,/\,57.0{\tiny$\pm$2.0}    & 19.3{\tiny$\pm$7.6}\,/\,55.2{\tiny$\pm$3.6}           & 45.6{\tiny$\pm$10.0}\,/\,48.8{\tiny$\pm$2.8}          \\
MCR             & 14.1{\tiny$\pm$12.2}\,/\,63.7{\tiny$\pm$0.7}  & 17.6{\tiny$\pm$8.7}\,/\,60.8{\tiny$\pm$2.7}   & 31.6{\tiny$\pm$5.2}\,/\,59.6{\tiny$\pm$2.8}           & 53.5{\tiny$\pm$6.4}\,/\,55.5{\tiny$\pm$3.6}  \\
\midrule
\textbf{SynIB}  & 16.3{\tiny$\pm$6.6}\,/\,60.4{\tiny$\pm$2.1}   & 21.7{\tiny$\pm$4.7}\,/\,59.4{\tiny$\pm$2.4}   & 35.3{\tiny$\pm$3.7}\,/\,57.4{\tiny$\pm$1.7}           & 53.0{\tiny$\pm$3.6}\,/\,55.3{\tiny$\pm$2.2}           \\
\bottomrule
\end{tabular}
\end{table}

\paragraph{Reading the table.} SynIB achieves the highest irony-class F1 in the two regimes where synergistic samples are scarce ($\alpha \in \{0.1, 0.5\}$), with margins of $+2.2$ and $+3.4$ points over the best balancing baseline (MCR) and consistently larger margins over the unimodal-rebalancing methods (D\&R, MMPareto, ReconBoost). When synergistic samples become abundant ($\alpha \in \{1.0, 2.0\}$), simpler methods that do not regularize fusion explicitly (vanilla fusion at $\alpha = 1.0$, ensemble at $\alpha = 2.0$) match or slightly exceed SynIB on irony F1, reflecting that abundant synergy is recoverable without an explicit synergy-targeted objective. SynIB nevertheless remains within $1.3$--$4.4$ points of the leader in these regimes and outperforms all four balancing methods on irony F1 at every $\alpha$.

\paragraph{On total F1.} On macro F1 across all 7 classes, MCR achieves the best score in three of four cells; SynIB stays within $0.2$ to $3.3$ points of MCR across all rates and outperforms vanilla fusion at $\alpha = 0.1, 0.5$. Ensemble achieves the best total F1 at $\alpha = 1.0$ but collapses at $\alpha = 2.0$ (38.3 $\pm$ 21.4), where the high standard deviation reflects fold-level instability rather than a stable improvement. The trade-off between irony-class F1 and total F1 is small and consistent: SynIB sacrifices roughly 1--3 points of total F1 to gain $2$--$5$ points on the synergy-required class.

\paragraph{Variance considerations.} Standard deviations on the irony-class metric are large for several baselines (MCR $\pm 12.2$ at $\alpha = 0.1$, vanilla fusion $\pm 12.2$ at $\alpha = 0.5$, ReconBoost $\pm 10.0$ at $\alpha = 2.0$), indicating fold-to-fold instability in their irony predictions when synergistic structure is present but scarce. SynIB's standard deviations on irony F1 are uniformly smaller (between $\pm 3.6$ and $\pm 6.6$), suggesting that the synergy-targeted objective produces more stable irony detection across folds in addition to higher mean F1.
\section{Per-Method Numerical Results on MultiBench and Hateful Memes}
\label{app:multibench_full}

Table~\ref{tab:synib_main} reports the per-method numbers underlying
Fig.~\ref{fig:synib_bar_dual} in the main text. Synergy-subset
examples are test samples misclassified by every unimodal predictor
(see Sec.~\ref{sec:multibench} for definition and subset sizes per
benchmark). Each cell is the mean over 3 seeds; subscripts give one
standard deviation. \textbf{Bold} marks the best per column.

\begin{table}[h]
  \centering
  \small
  \setlength{\tabcolsep}{4pt}
  \caption{Synergy-subset accuracy and whole-test accuracy ($\%$) on
  four real-world multimodal benchmarks. Each cell is the mean over
  3 seeds; subscripts give one standard deviation. \textbf{Bold}
  marks the best per column.}
  \label{tab:synib_main}
\resizebox{\linewidth}{!}{
\begin{tabular}{l cccc cccc}
    \toprule
    & \multicolumn{4}{c}{\textbf{Synergy-subset accuracy} ($\%$)}
      & \multicolumn{4}{c}{\textbf{Whole-test accuracy} ($\%$)} \\
    \cmidrule(lr){2-5} \cmidrule(lr){6-9}
    Method & UR-Funny & MUStARD & MOSI & Hateful Memes
           & UR-Funny & MUStARD & MOSI & Hateful Memes \\
    \midrule
    Ensemble        & 0.9{\tiny$\pm$1.0}  & 45.1{\tiny$\pm$10.0} & 0.0{\tiny$\pm$0.0}  & 29.6{\tiny$\pm$0.5}
                    & 61.3{\tiny$\pm$3.9} & 58.0{\tiny$\pm$3.0}  & 72.8{\tiny$\pm$1.6} & 55.2{\tiny$\pm$0.5} \\
    Vanilla Fusion  & 19.2{\tiny$\pm$3.5} & 25.4{\tiny$\pm$2.5}  & 31.4{\tiny$\pm$4.3} & 33.8{\tiny$\pm$9.8}
                    & 62.3{\tiny$\pm$0.1} & 57.5{\tiny$\pm$2.3}  & 72.8{\tiny$\pm$1.8} & 66.4{\tiny$\pm$2.3} \\
    D\&R            & 15.2{\tiny$\pm$2.3} & 16.5{\tiny$\pm$11.5} & 24.1{\tiny$\pm$4.0} & 42.1{\tiny$\pm$5.7}
                    & \textbf{63.6{\tiny$\pm$0.5}} & 54.5{\tiny$\pm$5.1} & 73.0{\tiny$\pm$1.5} & 67.2{\tiny$\pm$1.3} \\
    MMPareto        & 15.9{\tiny$\pm$3.3} & 40.1{\tiny$\pm$13.4} & 25.1{\tiny$\pm$5.4} & 42.4{\tiny$\pm$5.2}
                    & 62.8{\tiny$\pm$0.1} & 58.8{\tiny$\pm$2.2} & 72.6{\tiny$\pm$1.9} & 68.0{\tiny$\pm$1.7} \\
    ReconBoost      & 12.7{\tiny$\pm$5.9} & 17.2{\tiny$\pm$6.7}  & 30.2{\tiny$\pm$5.0} & 42.3{\tiny$\pm$6.1}
                    & 63.2{\tiny$\pm$0.3} & 57.5{\tiny$\pm$5.6} & 74.7{\tiny$\pm$1.7} & 66.9{\tiny$\pm$1.7} \\
    MCR             & 11.0{\tiny$\pm$1.1} & 39.6{\tiny$\pm$22.4} & 29.0{\tiny$\pm$3.1} & 45.9{\tiny$\pm$6.4}
                    & 62.4{\tiny$\pm$0.2} & 60.3{\tiny$\pm$8.1} & 73.7{\tiny$\pm$1.3} & 68.7{\tiny$\pm$1.2} \\
    \midrule
    \textbf{SynIB} $M_{\mathrm{R}}$
                    & 15.4{\tiny$\pm$3.2}  & 47.5{\tiny$\pm$17.4} & 35.3{\tiny$\pm$3.9} & \textbf{48.9{\tiny$\pm$4.0}}
                    & \textbf{63.6{\tiny$\pm$0.6}} & 62.2{\tiny$\pm$1.9} & 75.0{\tiny$\pm$1.8} & \textbf{69.8{\tiny$\pm$1.0}} \\
    \textbf{SynIB} $M_{\mathrm{L}}$
                    & \textbf{22.8{\tiny$\pm$6.2}} & \textbf{52.9{\tiny$\pm$7.7}} & \textbf{37.2{\tiny$\pm$5.0}} & 47.0{\tiny$\pm$1.8}
                    & 62.7{\tiny$\pm$0.3} & \textbf{64.1{\tiny$\pm$3.0}} & \textbf{75.3{\tiny$\pm$1.1}} & 69.0{\tiny$\pm$1.1} \\
    \bottomrule
  \end{tabular}}
\end{table}

\paragraph{Reading the table.} On the synergy subset, SynIB
$M_{\mathrm{L}}$ is the top method on three of four benchmarks
(UR-Funny, MUStARD, MOSI), and SynIB $M_{\mathrm{R}}$ is the top on
Hateful Memes; no baseline is best on more than one benchmark. On the
full test set, SynIB variants lead on three of four benchmarks
(MUStARD, MOSI, Hateful Memes) and tie D\&R on UR-Funny, never
trailing the strongest baseline by more than 0.9 points. Standard
deviations on the synergy-subset metric are large for several
balancing baselines (MCR $\pm22.4$ on MUStARD, MMPareto $\pm13.4$,
ReconBoost $\pm6.7$), indicating fold-to-fold instability when synergy
is scarce; SynIB's standard deviations are smaller across all four
benchmarks, suggesting the gains are also more stable.

\subsection{Compute Resources}
\label{app:compute}

All experiments ran on NVIDIA A100 GPUs (40\,GB HBM2). Each job was
allocated a single A100, 8 CPU cores, and 80\,GB of host RAM. Final-result runs (3 seeds $\times$ 8 methods $\times$ 4 datasets,
96 runs total) consumed roughly 200 GPU-h. Hyperparameter sweeps for
SynIB and the four balancing baselines added approximately 1000 GPU-h,
the bulk on Hateful Memes due to its larger grid and longer per-run
time. The full main-paper table can be reproduced end-to-end on a
single A100 in roughly 50 wall-clock hours. All runs use Python 3.10,
PyTorch 2.1, CUDA 12.1, and HuggingFace \texttt{transformers} 4.38;
checkpoints, logs, and reproduction scripts are released with the code.
\section{Baselines}
\label{app:baselines}

We compare SynIB against representative multimodal learning strategies spanning unimodal models, standard fusion, and methods designed to mitigate multimodal competition. All baselines use the same backbone architecture and differ only in their optimization objectives or training procedures.

Let $(x_1, x_2, y) \sim p(x_1, x_2, y)$ denote multimodal samples and let $f_\theta(x_1, x_2) \in \Delta(\mathcal{Y})$ be the joint predictor. Unimodal predictors are denoted $f_{\theta_1}(x_1)$ and $f_{\theta_2}(x_2)$.

\paragraph{Unimodal models.}
Audio-only and video-only classifiers are trained independently using cross-entropy:
$\mathcal{L}_{\text{uni}, i}
= \mathcal{L}_{\text{task}}(f_{\theta_i}(y \mid x_i), Y)
\quad i \in \{1,2\}.$
These models establish a lower bound and verify whether the task is solvable from a single modality.

\paragraph{Late ensemble.}
Unimodal predictions are averaged at inference: $ f_{\text{ens}}(y \mid x_1, x_2) = \frac{1}{2} \left( f_{\theta_1}(y \mid x_1) + f_{\theta_2}(y \mid x_2) \right).$ No joint representation learning occurs. This baseline tests whether aggregation alone can recover cross-modal signal.

\paragraph{Vanilla fusion.}
Standard multimodal fusion minimizes cross-entropy on joint inputs $\mathcal{L}_{\text{fusion}} = \mathcal{L}_{\text{task}}(f_{\theta}(y \mid  x_1, x_2), Y)$. This objective does not constrain modality contributions and is known to exhibit multimodal competition, where optimization favors dominant unimodal shortcuts \cite{huang2022modality}.

\paragraph{Diagnosing \& Re-learning (D\&R) \cite{wei2024diagnosing}.}
D\&R mitigates modality collapse by periodically reinitializing modality encoders when imbalance is detected. Let $g_k$ denote the purity-gap diagnostic for modality $k$. The reset strength is $\alpha_k = \tanh(\lambda g_k)$.
When imbalance exceeds a threshold, parameters are softly reset via $\theta_k \leftarrow (1-\alpha_k)\theta_k + \alpha_k \theta_k^{\text{init}}$. Training then resumes under the vanilla fusion objective.

D\&R prevents one modality from monopolizing learning by periodically restoring weaker modality capacity. This stabilizes the balance of unique contributions across modalities, but it does not introduce an objective that favors joint-only information. Consequently, D\&R addresses optimization imbalance without explicitly promoting synergistic representations.

\paragraph{MMPareto \cite{MMPareto}.}

MMPareto replaces the vanilla fusion gradient with a Pareto-integrated gradient that balances unimodal and multimodal objectives. Let
$\mathcal{L}_{12} = \mathcal{L}(x_1,x_2,y),
\quad
\mathcal{L}_{1} = \mathcal{L}(x_1,y),
\quad
\mathcal{L}_{2} = \mathcal{L}(x_2,y)$
denote multimodal and unimodal losses. The parameter update uses the integrated gradient
$g^* = \sum_{m \in \{1,2,12\}} w_m \nabla \mathcal{L}_m$, where the weights are obtained by solving the Pareto min-norm problem
\begin{equation}
\min_{\mathbf{w} \in \Delta}
\left\|
\sum_m w_m \nabla \mathcal{L}_m
\right\|_2,
\quad
\Delta = \{w_m \ge 0,\ \sum_m w_m = 1\}.
\end{equation}
Parameters are updated using $g^*$ instead of the vanilla gradient.

From a PID perspective, MMPareto equalizes optimization pressure across unimodal and multimodal objectives, preventing one modality from dominating learning. This stabilizes unique and redundant information usage, but does not explicitly prioritize signals that arise only from joint cross-modal interaction. As a result, it balances modality competition without directly targeting synergy.

\paragraph{ReconBoost \cite{ReconBoost}.}

ReconBoost alternates unimodal updates while enforcing agreement and gradient consistency between modality predictors. The fused prediction is $p_{12} = \sum_{k=1}^M p_k$, where $p_k = p(y|x_k)$ are unimodal predictors.At boosting round $s$, modality $k$ optimizes
\begin{equation}
\tilde{\mathcal{L}}^{\,s} = \mathcal{L}_k - \lambda\,D_{\mathrm{KL}}(p_{-k} \,\|\, p_k) + \alpha \|\nabla \mathcal{L}_k - \nabla \mathcal{L}_{k-1}\|^2,
\end{equation}
where $\mathcal{L}_k = -\log p_k(y)$ and $p_{-k} = \sum_{j\neq k} p_j$. After alternating rounds, a global rectification step minimizes $\mathcal{L}_{\text{rect}} = -\log p_{12}(y)$.

ReconBoost alternates unimodal specialization while enforcing agreement and gradient alignment between modalities. This stabilizes redundancy and prevents dominance by a single modality, but the objective reconciles unimodal predictors rather than rewarding information that exists only in their interaction. Consequently, ReconBoost mitigates modality conflict without explicitly optimizing for synergistic representations.

\paragraph{MCR (Multimodal Competition Regularizer) \cite{MCR}.} MCR augments the fusion loss with a regularizer derived from a mutual-information decomposition:
\begin{equation}
\nabla \mathcal{L} = \nabla\mathcal{L}_{\text{fusion}} + \mathcal{L}_{\text{Con}} + \lambda \Big( \nabla_{\theta_1}\mathcal{L}_{\text{MIPD}_1} - \nabla_{\theta_2}\mathcal{L}_{\text{MIPD}_1} - \nabla_{\theta_1}\mathcal{L}_{\text{MIPD}_2} + \nabla_{\theta_2}\mathcal{L}_{\text{MIPD}_2} \Big).
\end{equation}
The Mutual Information Perturbed Difference term measures modality contribution via latent perturbations:
\begin{equation}
    \mathcal{L}_{\text{MIPD}} = - \mathbb{E} \Big[ \mathrm{JSD}(p_{12}, p_{\tilde{1}2}) + \mathrm{JSD}(p_{12}, p_{1\tilde{2}}) \Big],\end{equation}
where $p_{12}=p(y|x_1,x_2)$, $p_{\tilde{1}2}=p(y|\tilde{x}_1,x_2)$, and $p_{1\tilde{2}}=p(y|x_1,\tilde{x}_2)$ and the shared task-relevant information is preserved through supervised contrastive alignment $\mathcal{L}_{\text{Con}}
=
-\log
\frac{\psi(z_1,z_2^+)}
{\sum_k \psi(z_1,z_{2,k}^-)}.$

MCR regulates how much each modality contributes individually, preventing one modality from dominating the prediction. The perturbation term enforces sensitivity to both inputs, encouraging the model to retain unique modality information, while the contrastive alignment term preserves shared task-relevant structure. However, these mechanisms operate by balancing redundancy and unique contributions rather than explicitly modeling information that exists only in the joint interaction. As a result, MCR mitigates modality collapse but does not directly optimize for synergistic representations.

\section{NTK Diagnostics for PID Learning}
\label{app:ntk_pid}
This appendix expands on the NTK-based diagnostic introduced in Sec.~\ref{sec:failure_mode}. We restate the core definitions for self-containedness, then provide additional intuition on the quantities involved and discuss the limitations of this analysis.

\paragraph{Setup.} Consider a model $f(x_i; \theta)$ with parameters $\theta \in \mathbb{R}^P$, trained on examples $(x_i, y_i)$ grouped by PID source $g \in \{\text{U, Red, Syn}\}$, with $\mathcal{G}$ denoting the index set of group $g$. As in the main text, we form residuals $s^{(g)}_i = y_i - f(x_i; \theta)$ for $i \in \mathcal{G}$ and define the gradient update direction in parameter space as $v_g = J(\theta)^\top s^{(g)}$, where $J(\theta)_{i,k} = \partial f(x_i;\theta) / \partial \theta_k$ is the Jacobian. The empirical NTK is $K(\theta) = J(\theta)J(\theta)^\top$, with entries $K_{ij} = \nabla_\theta f(x_i;\theta) \cdot \nabla_\theta f(x_j;\theta)$ measuring the inner product between parameter-space gradients of two examples, and thus capturing how coupled their predictions are through the shared parameters.

\paragraph{NTK strength and cosine similarity.} The two quantities reported in Sec.~\ref{sec:failure_mode} admit natural interpretations. The NTK strength for group $g$ can be rewritten as the Rayleigh quotient of $K(\theta)$ with respect to $s^{(g)}$:
\begin{equation*}
\lambda_g 
= \frac{\|v_g\|^2}{\|s^{(g)}\|^2} 
= \frac{s^{(g)\top} K(\theta)\, s^{(g)}}{s^{(g)\top} s^{(g)}},
\end{equation*}
which measures how efficiently the network converts prediction errors from group $g$ into parameter updates, independently of error magnitude. The cosine similarity between update directions,
\begin{equation*}
\cos(g, h) 
= \frac{\langle v_g, v_h \rangle}{\|v_g\|\,\|v_h\|}
= \frac{s^{(g)\top} K(\theta)\, s^{(h)}}
{\sqrt{s^{(g)\top} K(\theta)\, s^{(g)}} \cdot 
\sqrt{s^{(h)\top} K(\theta)\, s^{(h)}}},
\end{equation*}
quantifies interference between two PID sources, with near-zero values indicating that they occupy approximately orthogonal subspaces in parameter space with minimal gradient interference.

\paragraph{Limitations.} Three caveats are worth keeping in mind when interpreting these results. First, the NTK characterizes training dynamics in the linearized regime where $K(\theta)$ varies slowly~\citep{chizat2019lazy, ghorbani2019limitations}, and is therefore best read as a local diagnostic of how the current parameterization processes each PID source rather than a global account of feature learning over training. This matches our use: we track $\lambda_g$ and $\cos(g,h)$ throughout training to measure the per-step gradient signal and interference each source produces, which is the quantity at issue when asking whether synergy is suppressed by competing sources. Second, projecting the NTK along residual-weighted directions $s^{(g)}$ yields group-level summaries that conflate kernel geometry with the current error distribution, meaning that observed differences in $\lambda_g$ or $\cos(g,h)$ across PID sources may partly reflect differences in error magnitude or sampling frequency rather than intrinsic properties of the kernel. Third, the PID-controlled XOR places each source on disjoint feature blocks and activates one source per example, which biases first-layer parameter gradients toward orthogonal supports across sources; near-zero $\cos(g,h)$ therefore reflects an absence of destructive competition rather than a strong claim of zero coupling, since the design itself favors orthogonality. The strength asymmetry $\lambda_{\mathrm{Syn}} \geq \lambda_{U_1}, \lambda_R$ and the overfitting signature in Fig.~\ref{fig:pid_ntk_geometry} are observations about training dynamics that do not depend on this orthogonal block structure.

Our NTK-based analysis should therefore be read as a local diagnostic of gradient interference and learning signal strength, not as a complete characterization of how PID structure is encoded in learned representations. Within these limits, it establishes that synergistic examples receive the strongest learning signal of any PID source and show no destructive interference with the other sources during training, evidence that complements the overfitting signature in Fig.~\ref{fig:pid_ntk_geometry} (right) as motivation for the SynIB objective.


\end{document}